\documentclass[runningheads]{llncs}

\usepackage{eccv}
\usepackage[title]{appendix}

\usepackage{eccvabbrv}

\usepackage{graphicx}
\usepackage{booktabs}

\usepackage[accsupp]{axessibility}  

\usepackage{hyperref}

\usepackage{orcidlink}

\usepackage{xspace}
\newcommand\modelname{CIC-BART\xspace}
\newcommand\cocoent{COCO-Ent\xspace}
\newcommand\flickrent{Flickr-Ent\xspace}
\newcommand\cocoentssa{COCO-Ent-SSA\xspace}
\newcommand\flickrssa{Flickr-Ent-SSA\xspace}

\usepackage{algorithm}
\usepackage[noend]{algpseudocode}

\DeclareMathOperator{\sam}{sam}

\usepackage{wrapfig,booktabs}

\usepackage{xcolor,colortbl}
\usepackage{lipsum}

\begin{document}

\title{CIC-BART-SSA: Controllable Image Captioning with Structured Semantic Augmentation} 

\titlerunning{CIC-BART-SSA}

\author{Kalliopi Basioti\inst{1,2}\textsuperscript{*}
\and
Mohamed A. Abdelsalam\inst{2} \and
Federico Fancellu\inst{3}\textsuperscript{\dag}
\and
Vladimir Pavlovic\inst{1}\textsuperscript{\dag}
\and
Afsaneh Fazly\inst{2}
}

\authorrunning{K.~Basioti et al.}

\institute{
Rutgers University, New Jersey, USA
\email{\{kalliopi.basioti, vladimir\}@rutgers.edu} \and
Samsung AI Centre - Toronto, Toronto, Canada  \email{\{m.abdelsalam, a.fazly\}@samsung.com}\and
Solventum
\email{ffancellu@solventum.com}}

\maketitle
\begin{center}
    \centering
    \captionsetup{type=figure}
    \includegraphics[width=\textwidth]{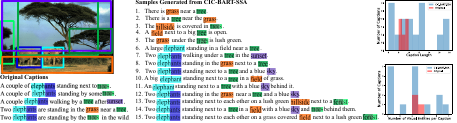}
    \captionof{figure}{Existing captioning datasets contain captions that describe the entirety of an image. This is reflected in the narrow distributions of the entities that appear in those captions and the caption lengths (the red-colored histograms). 
    CIC aims to generate diverse descriptions by controllably re-focusing on different spatiosemantic aspects of an image, such as the semantically coherent subsets of image objects. Our proposed \modelname-SSA is designed to produce diverse, controlled captions ranging from brief and concise to detailed and comprehensive. Sentences 1-15 are example outputs of our approach where the highlighted text indicates the focus of a controllable caption. The histograms demonstrate that our approach generates high-quality descriptions for a wider range of scene focus (number of visual entities) and caption length compared to the original captions. 
    \href{http://cocodataset.org/\#explore?id=108338}{Image} is licensed under \href{https://creativecommons.org/licenses/by-sa/2.0/}{CC BY-SA 2.0}.
    }
    \label{fig:motivation-diagram}
\end{center}
\begin{abstract}
{\let\thefootnote\relax\footnotetext{*Work done during an internship at Samsung AI Centre - Toronto}}
{\let\thefootnote\relax\footnotetext{{\dag}Work done while at Samsung AI Centre - Toronto}}Controllable Image Captioning (CIC) aims at generating natural language descriptions for an image, conditioned on information provided by end users, e.g., regions, entities or events of interest. However, available image--language datasets mainly contain captions that describe the entirety of an image, making them ineffective for training CIC models that can potentially attend to any subset of regions or relationships.  To tackle this challenge, we propose a novel, fully automatic method to sample additional focused and visually grounded captions using a unified structured semantic representation built on top of the existing set of captions associated with an image. We leverage Abstract Meaning Representation (AMR), a cross-lingual graph-based semantic formalism, to encode all possible spatio-semantic relations between entities, beyond the typical spatial-relations-only focus of current methods. We use this Structured Semantic Augmentation (SSA) framework to augment existing image--caption datasets with the grounded controlled captions, increasing their spatial and semantic diversity and focal coverage.  We then develop a new model, \modelname-SSA, specifically tailored for the CIC task, that sources its control signals from SSA-diversified datasets. 
We empirically show that, compared to SOTA CIC models, \modelname-SSA generates captions that are superior in diversity and text quality, are competitive in controllability, and, importantly, minimize the gap between broad and highly focused controlled captioning performance by efficiently generalizing to the challenging highly focused scenarios. Code is available at \url{https://github.com/SamsungLabs/CIC-BART-SSA}.
\end{abstract}    

\section{Introduction}
\label{sec:intro}

Image captioning refers to the task of providing an AI system with an input image, and asking the system to describe the visual content in natural language. This process requires the captioning system to understand what objects are present, in what context (e.g., event or scene), and how they relate. Recent deep learning approaches to this task \cite{rennie2017self, xu2015show, lu2018neural, cho2021unifying, li2020oscar, xia2021xgpt, ren2023crossing, ramos2023smallcap, luo2023semantic, zhong2020comprehensive} surpass human
performance in standard image captioning metrics. However, these models tend to generate general captions that describe the entirety of an image, and are often of limited diversity; see Original Captions in \cref{fig:motivation-diagram}. 
 
Controllable image captioning (CIC) overcomes these challenges by generating different descriptions for the same image in a user-controlled fashion. That is, a CIC model receives as input an image paired with a user-specified {\it control signal} (e.g., entities or regions of interest), and generates a caption conditioned on the control signal. CIC models are thus capable of generating a diverse set of captions by varying the control signal for the same image; see CIC generated captions $1$--$15$ in~\cref{fig:motivation-diagram}.

\begin{figure*}[t]
  \centering
    \includegraphics[width=\linewidth]{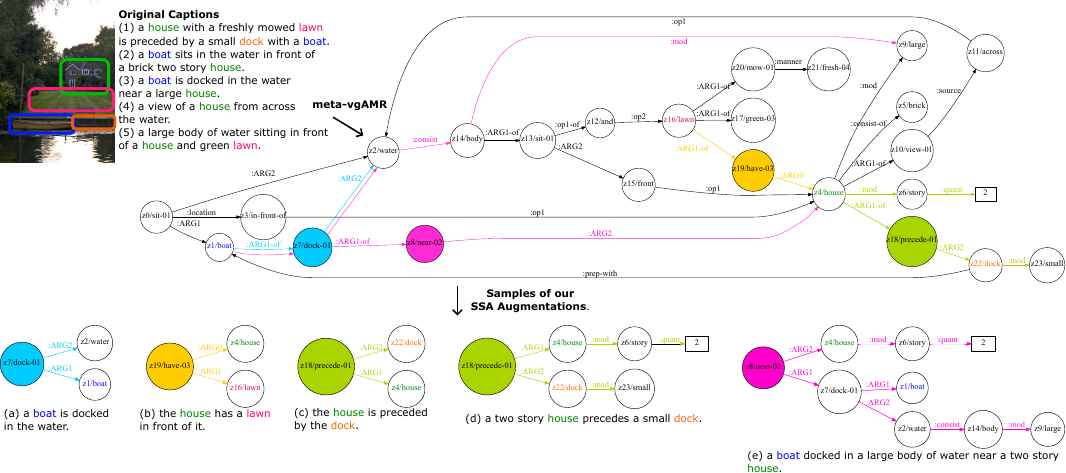}
    \caption{An example of our structured semantic augmentation approach. We start by using visually-grounded captions (1)-(5) to create a meta-vgAMR graph, which includes all available image information in one representation. We then sample sub-graphs from the meta-vgAMR to generate a new and diverse set of captions (such as sentences (a)-(e)). Our approach takes advantage of both linguistic and spatial diversity, with the latter creating descriptions for new combinations of visual entities. For instance, caption (a) focuses only on the `boat', and captions (c) and (d) focus on the `dock' and `house', combinations that are not explored in the original captions. 
    \href{https://farm3.staticflickr.com/2129/2432734812\_f1d31a8726\_z.jpg}{Image} is licensed under \href{https://creativecommons.org/licenses/by-sa/2.0/}{CC BY-SA 2.0}.
    }
    \label{fig:amr-motivation}
\end{figure*}

In realistic applications, the easiest way for the user to control the generation of captions is to limit the focus of the desired captions by selecting different entities (objects) using their bounding boxes, as shown in \cref{fig:motivation-diagram,fig:amr-motivation}.
Most previous work focuses on such spatial control signals \cite{johnson2016densecap, cornia2019show, wang2023caption, lindh2020language, zheng2019intention, zhao2022visual}. To improve performance, more recent studies supplement this spatial signal with additional information on the desired length, style, or syntactic and semantic structure of the generated text \cite{chen2020say,chen2021human}, increasing the richness and complexity of control signals.
However, for the CIC approach to succeed, the CIC models need to be trained on equally rich datasets that incorporate, explicitly or implicitly, those control signals. Unfortunately, most image captioning datasets today, such as Flickr30k~\cite{plummer2015flickr30k} or MS-COCO~\cite{cornia2019show}, lack this necessary diversity of controls and corresponding captions.   

Our goal is to achieve SOTA performance in CIC without the need for new, increasingly rich, yet also costly, and impractical-to-collect datasets, where human workers would face the burden of having to provide multitudes of control signals and corresponding descriptive captions.
To achieve this goal, we propose a novel Structured Semantic Augmentation (SSA) method, which automatically generates an augmented set of captions and the corresponding control signals with diverse spatiosemantic focus starting from only the core set of ``original'' uncontrolled captions.  The method takes advantage of a detailed visual-linguistic semantic graph (illustrated in~\cref{fig:amr-motivation}) constructed from the original captions and their image groundings. To build these semantic graphs, we use Abstract Meaning Representation (AMR)~\cite{banarescu2012abstract}, a semantic formalism that can capture fine-grained linguistic relations beyond the exclusively spatial relationships present in the common scene graphs~\cite{johnson2015image}. The availability of robust AMR parsers \cite{astudillo2020transition,blloshmi2021spring} allows us to generate semantic graphs for individual captions \textit{automatically}, which we then merge into a rich meta-AMR graph for the joint image--language pair. From this meta-graph, we sample diverse connected subgraphs that represent semantically coherent combinations of image-anchored entities, events, and their relations, which we then turn into controlled captions automatically via existing AMR-to-text models \cite{blloshmi2021spring}. \cref{fig:amr-motivation} depicts an example of our meta-graph inferred from the original uncontrolled captions associated with an image. Filled nodes in the meta-graph indicate image entity groundings.  Five semantically coherent subgraphs (a)--(e) of variable complexity are then sampled from the meta-graph, which are subsequently used to generate novel captions, shown below each subgraph. These new captions augment the original caption set by providing both image focus, through node groundings, and increased semantic diversity induced by the sampled subgraphs.
Building upon SSA, we introduce a new CIC model, \modelname, suitable for generating focused controlled captions. 
Alongside the regions of interest, \modelname also makes use of the length of the desired caption as a control signal proxy for the verbosity of the caption.  \modelname can be trained on SSA-augmented versions of standard VL datasets such as MS-COCO or Flicker30k to accommodate the CIC task.
Our experiments show that, compared to several SOTA models, the captions generated by our model have superior text quality and diversity, while being comparable in terms of faithfulness to control signals.

\noindent In summary, our contributions are:\\
\indent 1. We propose a novel data augmentation technique, SSA, that draws on a structured semantic formalism (AMR) to automatically generate focused captions suitable for training of CIC models. We empirically show that our SSA technique enables CIC models to generate captions with high controllability, diversity, and text quality. \\
\indent 2. We propose \modelname, a model designed for CIC, that does not require overly descriptive and complex control signals that SOTA models often require to achieve high performance.
We show a superior overall performance, compared to SOTA, while relying on simple control signals (i.e., regions of interest and preferred caption length). \\
\indent 3. We present an extensive evaluation of our model, compared with existing SOTA. Specifically, we report results on different aspects of generated captions, including controllability (faithfulness to control signal), diversity, and text quality (linguistic well-formedness). To account for the trade-off among these metrics, we propose an overall performance score based on their harmonic mean. This metric helps us identify models that perform well in all these aspects.

\section{Related Work}
\paragraph{Controllable Image Captioning (CIC).} Various types of control have been used for CIC, including visual entities, a type of region-based control \cite{johnson2016densecap, cornia2019show, wang2023caption, lindh2020language, zheng2019intention, zhao2022visual}, where generated captions should learn to focus on the regions of interest. Others draw on complex control signals where additional knowledge about the generated caption structure is provided.
For example, some recent work provides the complete skeleton of the desired sentence in the form of a number of objects or attributes or object-relation-object templates \cite{chen2020say, chen2021human}. 
Additional control signals that CIC draws on include different caption styles, e.g., positive, negative, humorous, or romantic tone \cite{wang2023caption, mathews2016senticap, zhao2020memcap, gan2017stylenet, guo2019mscap, mathews2018semstyle, wang2023controllable, zeng2023conzic}, user personality \cite{chunseong2017attend, shuster2019engaging}, or the
length of the generated captions \cite{deng2020length, wang2023caption, hirsch2022clid, wang2023controllable, wang2023learning}. 
The use of complex control signals aims at improving the diversity of captions and the quality of the text in CIC models. However, it requires the users to provide a detailed description of the control signal, which is not realistic in practical settings where such models are to be deployed (e.g., a self-driving car or personal assistant). We instead draw on two simple control signals (regions of interest and desired caption length) and show that we can achieve competitive performance on CIC, while keeping the control signals simple and practical. 

Recent SOTA models that draw on spatial control include the SCT model \cite{cornia2019show} that also uses the Faster R-CNN feature vectors and object tags (corresponding GloVe vectors \cite{pennington2014glove}) of the entities of interest, 
as well as models that include skeleton-based control, namely 
ASG2Caption \cite{chen2020say} 
and VSR \cite{chen2021human}.
ASG2Caption uses an abstract scene graph (ASG) to express the desired structure of a caption. 
ASG contains three types of unlabeled \textit{abstract nodes} (object, attribute, relationship) that are grounded in the image by extracting features from the corresponding bounding boxes (for objects and attributes) or from the union of bounding box pairs (for a relationship node).
ASG2Caption shows improved controllability (by conditioning on ASGs), and diversity (by automatically sampling diverse ASGs as control signals).
The VSR model \cite{chen2021human} draws on GloVe embeddings of Faster R-CNN object tags for visual entities (as in SCT). It also uses a skeleton control signal (like ASG2Caption), 
but one that includes more detailed information and richer semantics. Specifically, the VSR control signal follows the form of a fine-grained PropBank entry\footnote{\url{https://propbank.github.io}} --- i.e., specifying the exact verb(s) expressing action(s) depicted in the image, and their visually grounded arguments (e.g., subject, object, location, manner). 
Thus, VSR uses the most descriptive control signal among the SOTA models. Refer to \cref{sec:sup-setup-ssa-baseline} for an illustrative example of the control signal used for each method.

Compared to ASG2Caption and VSR, our control signal is kept minimal and only specifies the bounding boxes and desired caption lengths. To improve the diversity of captions, we draw on a structured semantic graph (AMR) that expresses the semantics of a sentence based on PropBank semantics. Notably, we do \textbf{not} use these rich graphs to express detailed and overly descriptive control signals (as in VSR), but we use these semantic structures to augment our training data with richer and diverse captions, which will result in the model learning to generate more diverse captions.   
Additionally, we include a length control signal to further increase diversity without needing to specify detailed information about the structure of the output (e.g., number of attributes per object, etc.). This way, we can generate a variety of captions for a fixed image sub-region by simply controlling the desired length of the output.

\paragraph{Abstract Meaning Representation (AMR).} 
AMR \cite{banarescu2013abstract} is a rich semantic formalism for expressing the meaning of natural language sentences as a formal graph. AMR draws on PropBank, which is a rich lexical semantic resource encoding predicates expressing an action or state, as well as the number and nature of the participating entities (arguments and other semantic roles, such as location, manner, etc.).
AMR is a widely researched semantic formalism for which highly accurate automatic Text-to-AMR and AMR-to-Text models are developed \cite{astudillo2020transition,blloshmi2021spring}. We rely on these models to augment original image--caption datasets with newly generated captions (as explained in Sect.~\ref{sec:ssa}).

\paragraph{AMRs vs. Scene Graphs.} Recent studies \cite{zellers2018neural, abdelsalam2022visual, choi2022scene, choi2022sgram} have shown that AMRs better capture the semantic relations of an image as compared to the scene graphs \cite{chang2021comprehensive}. Existing scene graph annotations mainly capture geometric or possessive relations, which account for more than 90\% of the relations captured, whereas more than $1/3$ of the captured entities refer to clothing, object, or body parts information \cite{zellers2018neural, abdelsalam2022visual}. This difference is crucial for high-quality image captioning, as we use higher-level semantic relations in our everyday language rather than geometric ones. For instance, during a soccer game, we would probably describe a goal save as `the player kicks the ball away from the goal' or `the goalkeeper defends his team by saving a goal' and not by using mainly geometric and possessive relations like `a person wearing a white shirt, standing with his right leg lifted, close to a ball which is above the ground'. In \cref{sec:sup-scenegraph-augmentations}, we provide a detailed comparison of AMR and scene graph representations, particularly focusing on their applications in data augmentation.
\section{Model}
\label{sec:model}
\begin{wrapfigure}{r}{0.5\textwidth}
\vspace{-1\baselineskip}
  \begin{center}
    \includegraphics[trim=0cm 9.5cm 3.5cm 0cm,clip=true,width=0.48\textwidth]{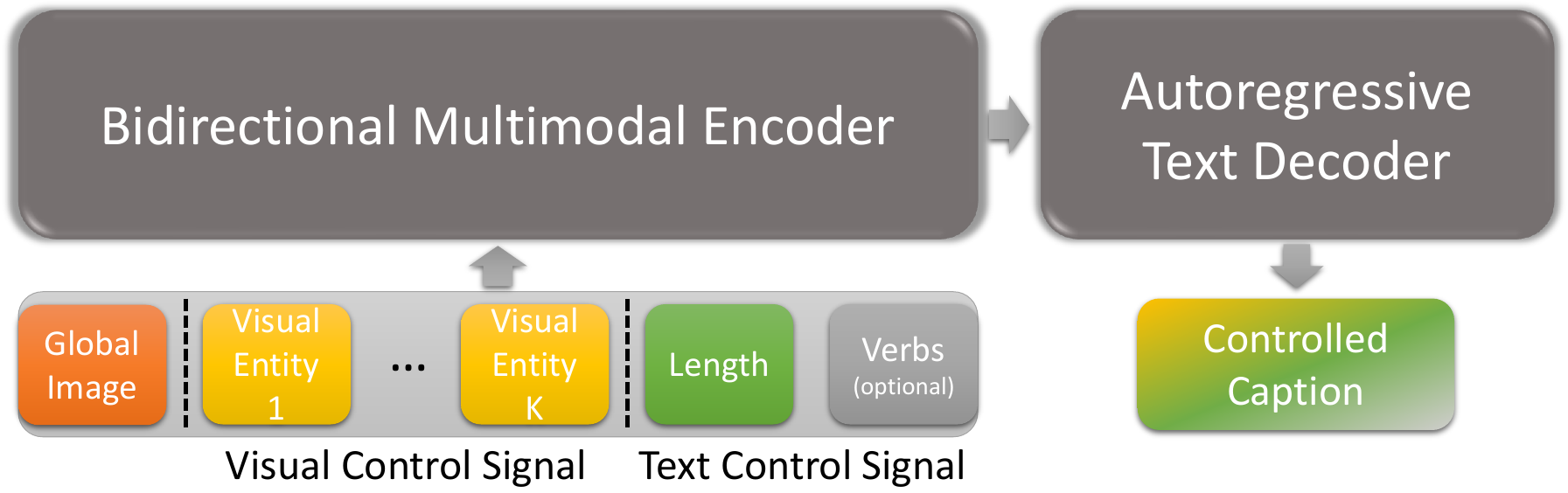}
  \end{center}
  \caption{
  The architecture diagram of our model, \modelname, which enables the generation of region- and length-controllable captions. When event information is available, the corresponding verb can be included in the control signal.
  }
  \label{fig:model-architecture}
\vspace{-1\baselineskip}
\end{wrapfigure}
We propose \modelname, specifically designed to generate controlled image captions. Specifically, it can generate descriptions of particular areas within a scene with a desired level of detail. Our model, based on VL-BART \cite{cho2021unifying}, utilizes a transformer-based encoder-decoder architecture, as shown in \cref{fig:model-architecture}. \modelname extends VL-BART encoder to the CIC task by modifying the encoder input to include:
a) a global image embedding that provides the context of the full image to the model;
b) the visual control signal, including the visual embedding of the regions that contain the entities of interest;
c) the text control signal, containing length control (indicating the desired length range of the output caption) and an \textit{optional} verb signal that indicates the action we want the generated caption to concentrate on.

The visual embeddings of the regions are position-aware embeddings from a Faster R-CNN model \cite{ren2015faster} trained for visual object and attribute classification \cite{Anderson2017up-down} on Visual Genome. The global image feature vector is extracted as well from Faster R-CNN. 
For the length control signal, we add to our vocabulary $L$ tokens for the $L$ different caption length levels; for instance, level one represents sentences between one and nine words, and level two, ten to nineteen. These tokens describe our coarse levels, for a finer sentence size accuracy, we accompany the tokens with the desired number of words. This choice gives our model the capacity to generate diverse captions for a particular length level. Finally, the output of the decoder generates the desired, controlled image caption.
\section{Structured Semantic Augmentation (SSA)}
\label{sec:ssa}

The goal of our SSA method is to augment existing image captioning datasets with new focused captions along with their control signals (i.e., regions corresponding to entities).
We rely on datasets where visual entities in the captions are annotated with their corresponding regions (see Sect.~\ref{sec:setup} for details on the datasets).
The SSA process consists of four main steps, as described below. For more details, refer to \cref{sec:supp-ssa}, which includes a step-by-step example of our SSA methodology.

\paragraph{Step 1: Image-level AMR graph generation.} Our objective in this stage is to enclose all the information available from the visually grounded captions into a single representation. To accomplish this, we create a visually grounded AMR graph (vgAMR) for each caption of an image and then merge them into a single image-level graph, the meta-vgAMR.
To create the vgAMRs of an image, we first convert each of its $N$ captions to their AMR representation, using the Neural transition-based Text-to-AMR parser \cite{astudillo2020transition} which also aligns words in a caption with their respective nodes in the AMR graph. We utilize the alignment information of 1) caption words and AMR nodes (from Text-to-AMR parser) and 2) caption words to image bounding boxes (from existing dataset annotations) to visually ground the AMR nodes. After this step, we get the collection of nodes referring to visual entities, where each grounded meta-AMR node is linked with a non-empty set of bounding boxes. This extended representation, `AMR + visual grounded nodes', is our vgAMR. 

Our next step is to combine the N vgAMRs to form a single meta-vgAMR. To achieve this, we employ a pairwise strategy to merge the most similar vgAMRs first (we measure similarity with Smatch score \cite{cai2013smatch}). We use the UPGMA hierarchical clustering algorithm \cite{lukasova1979hierarchical, mullner2011modern} to find the optimal merge ordering starting from the most similar graphs. UPGMA creates a hierarchy where the bottom level consists of the N individual vgAMRs. By merging all vgAMRs using the UPGMA ordering, we obtain a single structure called meta-vgAMR.

When merging two vgAMRs, the main challenge is identifying which nodes correspond to the same concepts, such as entities, attributes, actions, and relations. We use three node properties to accomplish this: a) visual grounding information, b) semantic similarity of node labels, and c) node neighborhood semantic similarity. We derive two node-merging criteria from there: 1) visually grounded entity nodes are merged if they point to the same image-bounding boxes. When 1) does not hold, we check the second criterion: 2) for the remaining non-grounded nodes, including amr-specific, predicates, adjectives, and adverbs, we use a combination of node label semantic similarity (cosine similarity of the labels using their GloVe embeddings) and neighborhood similarity. Neighborhood similarity examines the similarity of parents for adjectives/adverbs nodes and children for predicate nodes, along with the similarity of connecting edge roles. When two nodes satisfy criterion 1) or 2), we merge them into a single node. 
Moreover, if they have different labels, we maintain both names by keeping a list of synonyms to increase representation diversity. In \cref{sec:supp-ssa} (\cref{fig:flow-diagram-node-merge}), we have included the flow diagram depicting the process of merging two nodes corresponding to the same concepts.

In the special case when the two vgAMRs describe two totally different concepts, and hence they have no common nodes, we add an amr-specific node called `multi-sentence' as the root with the two independent vgAMRs as its children. 
The final graph, meta-vgAMR, includes all non-redundant\footnote{A node may have different names for the same bounding box in different meta-vgAMRs, such as `A male' and `A person'. According to criterion 1), we merge the corresponding AMR nodes and keep both 'male' and 'person' in the names list to avoid redundancy. Therefore, criteria 1) and 2) ensure that multiple nodes don't describe the same concept in the meta-vgAMR.} elements of the original $N$ captions while preserving the visual grounding between the meta nodes and their respective image regions. 

\noindent\underline{Remark:} Meta-vgAMR efficiently compresses all available image information into a single structure. Following our approach, we can easily scale when new scene information becomes available by applying our pairwise merge procedure.

\paragraph{Step 2: Event-based graph sampling from image-level AMRs.} We start from the predicate nodes, which mainly correspond to verbs, to sample subgraphs in meta-vgAMR graphs. Predicate nodes are identified by their label and the edges connected to them. The label of a predicate node typically follows the format `predicate\_name-xx,' where `xx' represents the different senses a word can have regarding the concept it is used for. Predicate nodes have outward ARGy edges, where `y' can take values from 0 to 5, connecting them to their arguments. 
We sample subgraphs from these nodes by following the outgoing argument edges, which are labeled as ARG\textit{n} in an AMR graph, each defining a particular semantic role (e.g., ARG0 points to the agent, ARG1 to the patient, etc.).
Finally, we add one more subgraph containing the remaining children branches of other non-ARG optional predicate edges (e.g., `location', `time'). We repeat this process until the leaves of the graphs are reached. 
During sampling, we randomly select one of the synonyms if a node is a list of synonym labels, as mentioned in the previous step. The output of this step is our more focused \textit{event-focused} sub-graphs. In \cref{fig:amr-motivation}, we can see some instances of our event-based sampling (SSA samples), where the predicate nodes include z0/sit-01, z7/dock-01, z13/sit-01, and so on\footnote{Note that in \cref{fig:amr-motivation}, the node z17/green-03 is also categorized as a predicate. This may seem an error because we usually think of `green' as an attribute node rather than a predicate. However, in AMRs, when `green' is paired with its argument, in this case, z16/lawn, it encapsulates a predicate/verb that can be expressed in natural language as `the lawn is green.'}. Although we cannot show all the sampled event-based subgraphs in the figure, we included five of them and used colored roots and edges for visualization purposes.

\paragraph{Step 3: New caption generation from sampled AMRs} We use the SPRING AMR-to-Text model \cite{bevilacqua2021one} to generate new event-focused captions from the sampled vgAMR subgraphs. Because both vgAMR merging and sampling steps introduce noise, the output captions are not always of good quality. We automatically filter low-quality captions by using a linguistic well-formedness measure, GRUEN \cite{zhu2020gruen}, which is a reference-free metric based on BERT contextual embeddings. 
In \cref{sec:sup-augmentations-ablation} we provide examples of original dataset captions and their SSA augmentations, along with their GRUEN score.  

\paragraph{Step 4: Control signal generation.} The last step is to create the control signal for the generated captions. 
The spatial control signal for a specific caption is extracted from the corresponding sampled vgAMR, by pulling the bounding boxes of the visual entity linked AMR nodes.

\subsection{Mixing Strategies of Original and SSA Data}
\label{sec:ssamix-strategy}

To analyze the impact of our SSA data, we explore various mixing strategies with the original training set. Assume ${\mathcal D}$ represents the training control-caption pairs in the original dataset, containing $N_\mathcal{D}$ samples, and $SSA$ represents our SSA samples, containing $N_{SSA}$ instances. The augmented dataset ${\mathcal D}_{SSA}$ is defined by combining ${\mathcal D}$ and $SSA$: ${\mathcal D}_{SSA} = \sam_{\mathcal{D}}(\tau_{\mathcal{D}}, p_{\mathcal{D}}) \cup \sam_{SSA}(\tau_{SSA},p_{SSA})$, where the functions $\sam_{\mathcal{D}}$ samples a subset of the original dataset, and $\sam_{SSA}$ a subset of our SSA data. Since we are interested in the effect of our SSA, we assume that $\sam_{\mathcal{D}}(\tau_{\mathcal{D}}, p_{\mathcal{D}}) = \mathcal{D}$, with $\tau_{\mathcal{D}}=\text{`Random Sampling Strategy'}$ and $p_{\mathcal{D}}=100\%$, meaning that all original data are included in the mixed dataset. Depending on the $\sam_{SSA}$ parameter $\tau_{SSA}$ we have the cases:

\paragraph{Random Sampling Strategy.} In this case, we randomly select a pre-specified number of examples from $SSA$. The parameter $p_{SSA}$ expresses the percentage of SSA samples included in ${\mathcal D}_{SSA}$. With boundary cases $p_{SSA}=100\%$ (all $N_{SSA}$ samples are included), and $p_{SSA}=0\%$ (no SSA data are added).

\paragraph{Uniform-Coverage Sampling Strategy.} To mitigate the original dataset's bias (having mainly samples describing the entire image), we aim to create a new focus-unbiased dataset. By modeling the control signal focus as the image area percentage covered by the bounding boxes of the control signal, we split the original data into $B$ coverage bins. Then, we will randomly add in each bin SSA samples, aiming to create a new uniform, coverage-unbiased ${\mathcal D}_{SSA}$ dataset. Here, $p_{SSA}$ contains the range of each bin for the coverage histogram. For example, in the case where we choose ten uniform coverage bins, we have $p_{SSA} = \{ [0\%,10\%), [10\%,20\%), \dotsc, [90\%,100\%] \}$.

We present results from the Random Sampling Strategy for $p_{SSA}=0\%$ and $p_{SSA}=100\%$ in the main paper. Results from other scenarios can be found in \cref{sec:supp-ssamix-strategy}.

\section{Experimental Setup}
\label{sec:setup}
\subsection{Data}
We use Flickr30k Entities (Flickr-Ent) \cite{plummer2015flickr30k} and 
MS-COCO Entities (COCO-Ent) \cite{cornia2019show} for training and evaluation.
Flickr-Ent augments the original captions of Flickr30k \cite{young2014image} with manually-annotated region--phrase groundings. 
Flickr-Ent contains the original $31K$ images annotated with five captions each.
COCO-Ent augments the original MS-COCO \cite{lin2014microsoft} ($120K$ images each annotated with around five captions) with semi-automatically collected grounding annotations; see \cite{cornia2019show} for details on the annotation process. 
For both datasets, we follow previous work and use the training and test splits by Karpathy \etal \cite{karpathy2015deep}. 
We apply our SSA algorithm on the aforementioned datasets to create their augmented variations, \cocoentssa\ and \flickrssa, containing about $800K$ and $250K$ training captions, respectively, of which $33\%$ and $37\%$, are generated by our SSA algorithm.

For all four training sets, we automatically generate image--control--caption triplets to train our model on. For spatial control, we extract from the grounded captions the bounding boxes of the entities of interest using the annotations from \cocoent and \flickrent.
Note that since these datasets do not contain the text control signal, we use each ground-truth caption as a proxy for a controlled caption, from which we first generate the coarse- and fine-length control levels and then extract their verbs using part-of-speech tagging for the optional action control. 

\subsection{Models and Evaluation Metrics} 

We compare two variations of our model (with and without SSA augmentations) with SOTA models as our baselines:
Show Control \& Tell (SCT) \cite{cornia2019show} that uses region-based control (bounding boxes of visual entities of interest); ASG2Caption (ASG) \cite{chen2020say} that draws on visually grounded abstract scene graphs as control signal; and VSR \cite{chen2021human} that uses overly descriptive control signals that express verb (s) and fine-grained verb-specific semantic roles of the desired captions; ComPro \cite{wang2023learning} that learns a mapping from the bounding boxes of the entities of interest and caption length level to GPT-2 Large prompts aiming to retrieve controlled captions; and the LaBERT length-control-only model \cite{deng2020length}. 

We report the performance of our models and baselines using a comprehensive set of metrics that evaluate different aspects of caption controllability, diversity, and quality. We also propose and report an overall performance metric that summarizes these different aspects in a meaningful way. 
In particular, for diversity, we measure n-gram diversity D-1, D-2 \cite{aneja2019sequential} and self-CIDEr (sC) \cite{wang2019describing} metrics. For content controllability, we have developed an extended version of the IoU \cite{cornia2019show} and further analyzed its performance by introducing the Hallucinating Nouns (Hal) metric. Both metrics are thoroughly discussed in \cref{sec:metrics-supp}. For length controllability, we measure mean absolute error (L) and length precision (LP) \cite{deng2020length}. We assess the generated text quality using the GRUEN (G) \cite{zhu2020gruen} metric. We determine the overall performance using the harmonic mean of IoU, G, and sC metrics. A higher score indicates better performance. The harmonic mean ($H$) helps us determine the model with the best overall performance since it prioritizes models that perform well across all metrics while penalizing those with poor performance, even in one metric.
\cref{sec:metrics-supp} provides details for each evaluation metric. Further, in \cref{sec:standard-captioning} we include comparisons of the CIC models on standard captioning metrics (like CIDEr \cite{vedantam2015cider} and Spice \cite{anderson2016spice}). Note that standard captioning metrics are not sufficient for evaluating CIC, as they compare a generated (controlled) caption with a reference ground-truth caption, ignoring the desired effect of the control signal. We report them for completeness, but we believe that our well-formedness metric (GRUEN) is more suited for evaluating the quality of the controlled captions.
\begin{table}[t!]
\caption{
Performance of CIC models based on content controllability (IoU), text quality (G), and diversity (sC, D-1, D-2), and their harmonic mean (H).
For our models, we also report length controllability (L); baseline models (SCT, ASG, VSR) do not include this type of control. 
All models are evaluated on the original \flickrent and \cocoent test sets.
$^*$\textit{ASG-type dataset is not released for \flickrent; therefore, we could not reproduce the ASG results. 
}
}
\scriptsize
\centering
\setlength{\tabcolsep}{2pt}
\begin{tabular}{@{}l>{\columncolor[gray]{.9}}ccccccc|>{\columncolor[gray]{.9}}ccccccc}
\toprule
Model         & $H\uparrow$ & IoU$\uparrow$ & G$\uparrow$ & sC$\uparrow$ & D-1$\uparrow$   & D-2$\uparrow$ & L$\downarrow$ & $H\uparrow$ &IoU$\uparrow$ & G$\uparrow$ & sC$\uparrow$ & D-1$\uparrow$   & D-2$\uparrow$ &  L$\downarrow$   \\
\midrule
&\multicolumn{7}{c}{\cocoent} & \multicolumn{7}{c}{\flickrent}                                              \\
SCT\cite{cornia2019show}    & 55.8           & 67.3             & 64.4             & 42.8            & 27.0           & 35.5           & -       &  54.6          & 50.7           & 79.8           & 44.0           & 29.3           & 36.5 & -       \\
ASG*\cite{chen2020say}      & 74.2           & 72.6             & 72.0             & 78.3            & 37.8           &\underline{56.6}& -       & - & -          &  -             & -              & -              & -              &-\\
VSR\cite{chen2021human}     & 56.2           & \bf{77.6}        & 39.0             & 67.4            & 30.0           & 42.2           & -       & 62.5           & \bf{60.2}      & 54.0           & 77.9           & 33.3           & 49.3 &-      \\
\modelname                  &\underline{75.9}& 76.2             & \underline{73.0} & \underline{78.7}&\underline{38.0}& 56.2           & .49     &\underline{69.8}& 54.0           &\underline{85.0}&\underline{78.6}&\underline{43.6}& \underline{58.2}            & 1.24 \\
\modelname-SSA              &\bf{78.3}       & \underline{77.2} & \bf{74.8}        & \bf{82.5}       & \bf{44.6}      & \bf{63.2}      & \bf{.11}& {\bf71.3}      &\underline{55.0}& \bf{86.0}      & \bf{81.7}      & \bf{47.0}      & \bf{62.6}       & \bf{1.05}        \\
\bottomrule
\end{tabular}
\label{tab:overall}
\end{table}
\begin{table}[t!]
\caption{
Performance of CIC models based on content controllability (IoU), text quality (G), and diversity (sC, D-1, D-2), and their harmonic mean (H).
All models are evaluated on the \flickrent and \cocoent SSA-only test set samples.
}
\scriptsize
\centering
\setlength{\tabcolsep}{3pt}
\begin{tabular}{@{}l>{\columncolor[gray]{.9}}cccccc|>{\columncolor[gray]{.9}}cccccc}
\toprule
Model         & $H\uparrow$ & IoU$\uparrow$ & G$\uparrow$ & sC$\uparrow$ & D-1$\uparrow$   & D-2$\uparrow$ & $H\uparrow$ &IoU$\uparrow$ & G$\uparrow$ & sC$\uparrow$ & D-1$\uparrow$   & D-2$\uparrow$   \\
\midrule
&\multicolumn{6}{c}{\cocoent (SSA only)} & \multicolumn{6}{c}{\flickrent (SSA only)}                                              \\
SCT\cite{cornia2019show}    & 51.7           & \underline{62.1}             & 64.8             & 37.8            & 23.7           & 31.0           &    43.9        & 29.9         & 77.3 & 45.7   & 31.0           & 36.7                   \\
\modelname                  &\underline{69.2}& 61.4             & \underline{73.9} & \underline{74.0}&\underline{44.2}& \underline{57.0}           &\underline{68.5}& \underline{53.0}           &\underline{80.5}&\underline{79.8}&\underline{52.9}& \underline{62.9}            \\
\modelname-SSA              &\bf{75.6}       & \bf{65.2} & \bf{80.7}        & \bf{83.7}       & \bf{53.8}      & \bf{67.8}      & {\bf72.0}      &\bf{55.6}& \bf{82.9}      & \bf{86.1}      & \bf{56.5}      & \bf{69.3}              \\
\bottomrule
\end{tabular}
\label{tab:overall-ssa-only}
\end{table}
\section{Results}
\subsection{Overall Performance}
We first compare the overall performance of our models with the SCT, ASG, and VSR baselines with respect to controllable captioning metrics. We do not include ComPro in this comparison due to the unavailability of the codebase. We also exclude LaBERT since this model solely focuses on length controllability.

\cref{tab:overall} presents results for content and length controllability (IoU and L), text quality (G), and diversity (sC, D-1, D-2), as well as the harmonic mean (H).
For both datasets, \modelname-SSA has the best performance in all metrics, except IoU, where it is the second best.
Specifically, \modelname-SSA is superior to all other models with respect to diversity (sC, D-1, D-2) and text quality (G), but comparable to VSR in terms of content controllability (IoU). The length controllability (L) scores show that our SSA augmentation helps the model learn to generate high-quality output at the desirable length (compare \modelname and \modelname-SSA). This is due to the increased diversity in caption length provided by our SSA augmentations.

Importantly, we can see that model performance can vary depending on the metric. E.g., whereas VSR has the highest IoU, it falls behind in text quality and diversity. In our qualitative analysis, we observe the poor quality of the captions generated by VSR. 
The best-performing model should be identified based on the $H$ score that summarizes content controllability, text quality, and diversity into a single score. Based on this score, \modelname-SSA is better than SCT and VSR baselines by a large margin, and notably better than ASG. Nevertheless, ASG requires complex control signals in the form of scene graphs, in contrast to the simple control signal requirements of \modelname.

Next, we conduct a further evaluation of the CIC performance on our SSA samples from the test set images of \cocoent and \flickrent. We present the results in \cref{tab:overall-ssa-only} for our models \modelname, \modelname-SSA, and SCT. ASG and VSR are excluded since they need complex control signals (grounded abstract scene graphs for ASG and grounded verb semantic roles for VSR), which are only available for \cocoent and \flickrent. We observe a significant improvement in overall performance for our \modelname-SSA model. We notice that the models (SCT and \modelname) trained on the original datasets, which describe the entire image, had difficulties generalizing to cases where they had to focus on a specific sub-region of an image. However, our model \modelname-SSA was able to generate focused and diverse descriptions of the challenging, highly focused examples present in our SSA data.

\begin{wraptable}{r}{5cm}
\vspace{-1\baselineskip}
\caption{Length Precision (LP) for CIC models on \cocoent and \flickrent original test sets.}
\scriptsize
\label{tab:length-precision}
\centering
\begin{tabular}{@{}lc|c}
\toprule
Model & LP$\uparrow$ & LP$\uparrow$ \\
\midrule
&\cocoent & \flickrent  \\
ComPro\cite{wang2023learning}     & 94.7 & 81.4\\
LaBERT\cite{deng2020length}       & \underline{99.7} & \bf{98.4}\\
\modelname                        & \bf{99.9} & 88.0\\
\modelname-SSA                    & \bf{99.9} & \underline{91.3}\\
\bottomrule
\end{tabular}
\vspace{-1\baselineskip}
\end{wraptable}

We compare the length precision of our model with the baselines utilizing length control in Table \ref{tab:length-precision}. LaBERT uses only length-control signals without spatial control, while our model employs both spatial and length-control signals to generate focused captions. This makes the LaBERT task much easier since it only focuses on generating specific length descriptions of an image. On the other hand, our model focuses on generating captions that describe only a specific sub-region of the scene while maintaining a desired description length level. Although our task is more challenging than LaBERT, we achieve competitive length precision performance. 
Lastly, we want to emphasize that the improvement in length controllability (L) and length precision (LP) from \modelname to \modelname-SSA stems from the increased length diversity found in our SSA augmentations, which enriches the original \cocoent and \flickrent datasets. 
In \cref{sec:datastats}, we provide an analysis of the caption length statistics in the original datasets and our SSA-derived captions. 

\subsection{Effect of SSA on Content Controllability}
\begin{figure}[ht]
    \centering
    \includegraphics[width=\linewidth]{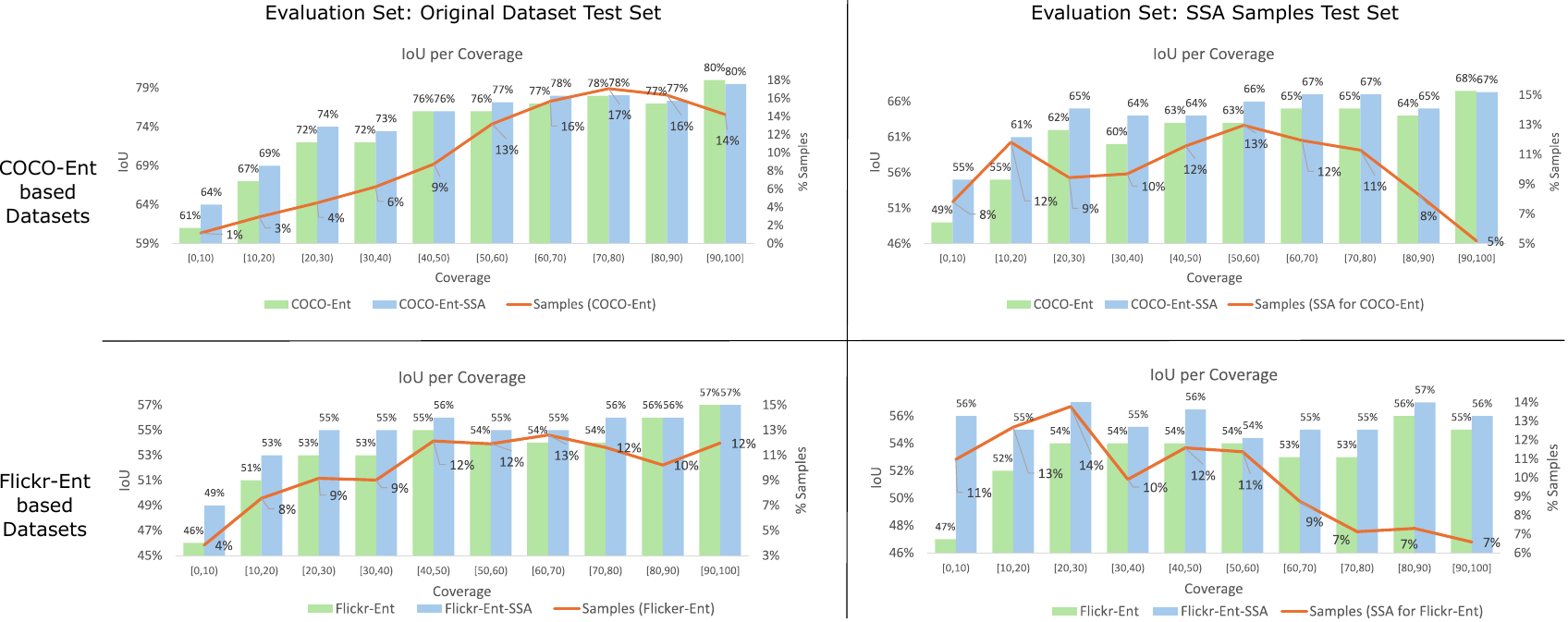}
    \caption{
    Content controllability (IoU) performance of \modelname when trained with (blue) and without SSA (green). The first column depicts the IoU on the original test set, and the second on the original test set images' SSA (only) data. The abscissa of each bar plot is the \% of the image covered by the control signal, so the left and right parts of the graph represent more focused and broader control signals, respectively. The \%Samples curve (orange) represents the distribution of test images in each coverage interval. The results show that SSA plays a crucial role in boosting \modelname performance in data-deprived, focused CIC settings.
}
    \label{fig:coverage-analysis}
\end{figure}

To analyze the impact of our SSA augmentations, we measure the content controllability (IoU) performance of \modelname at different levels of focus of the control signals and report it in \cref{fig:coverage-analysis}. We use coverage, defined as the area of the image enclosed by the bounding boxes of the entities of interest in the control signal, to quantify that focus. For example, highly focused control signals cover a small area, yielding low coverage, while broader signals cover a larger area and have high coverage. We `break down' the IoU performance into 10 coverage bands and report the average IoU over control signals in those bands. In addition, the `Samples' curve shows the distribution of test captions over the same bands. 
The results in \cref{fig:coverage-analysis} indicate that by training with SSA (blue bars), the spatial controllability improves significantly in the low-coverage regime, where the control signals are highly focused. Interestingly, these are also the most underrepresented (data deprived) parts of the original dataset \flickrent. Therefore, SSA which enriches the original datasets with highly focused examples (refer to \cref{sec:datastats} for \% Samples per coverage bands for the training sets), is effective in improving generalization performance in CIC.

\begin{figure}[t!]
  \centering
    \includegraphics[width=\linewidth]{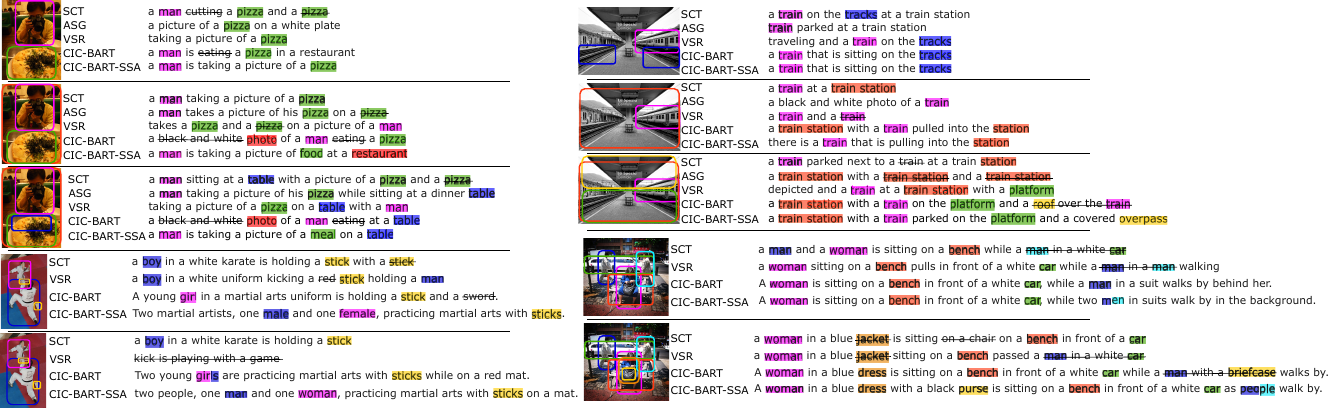}
  \caption{Qualitative examples for the original test sets. Strikethrough marks hallucinations and redundancies. \href{https://farm5.staticflickr.com/4102/4888234256_538b8dee56_z.jpg}{a}, \href{https://farm9.staticflickr.com/8501/8308004994_44eb2d562d_z.jpg}{b} licensed under \href{https://creativecommons.org/licenses/by-sa/2.0/}{CC BY-SA 2.0}; \href{https://flickr.com/photo.gne?id=101362133}{c}, \href{https://www.flickr.com/photos/moriza/151970521/}{d} under \href{https://creativecommons.org/licenses/by/2.0/}{CC BY 2.0}.
  }
\vspace{-1\baselineskip}
  \label{fig:qual}
\end{figure}

\subsection{Qualitative Analysis}
\begin{wrapfigure}{r}{0.5\textwidth}
\vspace{-2.5\baselineskip}
  \begin{center}
    \includegraphics[width=0.48\textwidth]{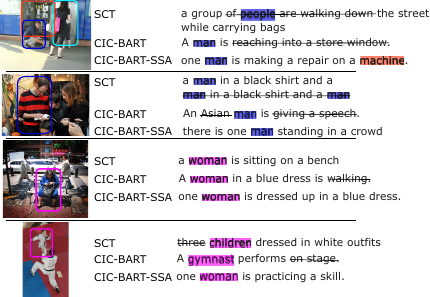}
  \end{center}
  \caption{Examples for SSA-only test set. Strikethrough marks hallucinations.
  Up to Down: \href{https://www.flickr.com/photos/thunderchild5/183647966/}{a} licensed under \href{https://creativecommons.org/licenses/by/2.0/}{CC BY 2.0}; \href{https://flickr.com/photo.gne?id=7249763658}{b} under \href{https://creativecommons.org/publicdomain/mark/1.0/}{CC PDM 1.0}, for last two see \cref{fig:qual}.
  }
  \label{fig:qual-ssa-main}
\vspace{-2\baselineskip}
\end{wrapfigure}

In \cref{fig:qual}, we present qualitative examples from the original test sets, and in \cref{fig:qual-ssa-main} examples from our SSA (only) test set control signals. In the two figures, each highlighted word found in the generated controlled captions corresponds to the control entity of the same color. This shows the match between the captions produced and the control signal. We also strike through the parts where the model hallucinates or generates redundant references to the entities of interest. 

Our models have been observed to outperform the previous state-of-the-art models by substantially enhancing the quality of the generated controlled captions. This behavior was expected from our quantitative analysis, which showed that our models have significantly higher text quality (G). More importantly, our CIC-BART-SSA model is capable of generating captions that are faithful to the control signal and better understand the relationships that connect the entities of interest.  
We include additional qualitative samples in \cref{sec:supp-qualresults}.

\section{Conclusions}

We address two main challenges faced by the controllable image captioning (CIC) models. First, standard image--caption datasets lack the controllability and diversity needed for proper training and evaluation of CIC. Second, most recent SOTA models require complex and overly descriptive control signals as input (including, e.g., the main action/verb to appear in the generated caption).
To address the first challenge, 
we propose a novel technique that draws on a structured semantic augmentation (SSA) formalism to generate focused captions and the corresponding control signals for images.
For the second challenge, we propose a transformer-based vision-language model attuned to the CIC task. We show that this model performs competitively with SOTA models without requiring complex and explicit control signals. 
Importantly, when combined with our SSA approach, our model generates highly diverse captions and significantly reduces the content controllability performance gap between the different levels of focus of the generated controlled captions. Finally, when provided with the commonly used verb guidance of other SOTA approaches, our model shows a substantial improvement in performance.
%
%
\bibliographystyle{splncs04}
\bibliography{main}
\clearpage
\appendix

\section{Overview}
\label{sec:supp}
In the appendix, we provide more details for our SSA methodology in \cref{sec:supp-ssa}; in \cref{sec:exp-setup-supp}, we provide additional details for our experimental set-up; and finally, in \cref{sec:result-supp} we provide extended qualitative and quantitative results of our performed experiments and ablations. Specifically, we focus on:
\begin{itemize}
    \item Dataset statistics before and after SSA augmentation in \cref{sec:datastats}.
    \item Impact of mixing of original and SSA captions in \cref{sec:supp-ssamix-strategy}.
    \item Effects of SSA on content controllability in \cref{sec:ssaciceffects}.
    \item SSA-induced diversity in \cref{sec:diversity}.
    \item Standard Captioning Performance of CIC models \cref{sec:standard-captioning}
    \item Qualitative comparisons in \cref{sec:supp-qualresults}.
    \item Comparison of SSA and alternative augmentation strategies with attention on LLM-based paraphrasing and Scene Graph-based methods in \cref{sec:sup-augmentations-ablation}.
\end{itemize}

\begin{algorithm}[t!]
\caption{\textbf{meta-vgAMR Graph Construction}} \scriptsize
\algrenewcommand\algorithmicindent{1em}
\begin{algorithmic}[1]
\footnotesize
    \State {\bf Input}: An image $I$ with $N$ human-generated, visually-grounded captions; We denote the visually grounded entities of each caption as $\{G^{en}_i\}_{i=1}^N$;
    \State {\bf Output}: The meta-vgAMR graph, $\mathcal{A}^{vg}_{Meta}$, of the $N$ captions;
    \State {\bf Initialize}: Generate the individual AMR graphs $\{\mathcal{A}_i\}_{i=1}^N$ for each image caption using a pre-trained Text-to-AMR semantic parser with (AMR node--caption word) alignment; Construct the vgAMRs, $\mathcal{A}^{vg} = \{\mathcal{A}^{vg}_i\}_{i=1}^N $, using the visual grounding annotations and (AMR node--caption word) alignment;
    \State Compute $D = 1 - $ SmatchScore$(\mathcal{A}^{vg})$; \Comment{\textcolor{blue}{A symmetric $N\times N$ AMR graph distance matrix between all $\mathcal{A}^{vg}_i,\mathcal{A}^{vg}_j$ pairs.}}
    \State bottomUpHCs = UPGMA$(D)$; \Comment{\textcolor{blue}{Bottom-up hierarchical clusters, each cluster contains two vgAMR graphs.}}
    \For {$(\mathcal{A}^{vg}_i,\mathcal{A}^{vg}_j)$ in bottomUpHCs} \Comment{\textcolor{blue}{Following the bottom-up hierarchy, pair-wise merge the vgAMRs of each cluster.}}
        \State $\mathcal{A}^{vg}_i = (\mathcal{N}_i, \mathcal{E}_i)$; $\mathcal{A}^{vg}_j = (\mathcal{N}_j, \mathcal{E}_j)$ \Comment{\textcolor{blue}{The nodes and edges of each vgAMR graph.}}
        \State Initialize $\mathcal{A}^{vg}_m = (\mathcal{N}_m, \mathcal{E}_m)$ as a null graph;
        \State $\mathcal{N}_{\text{common}}$ = getCommonNodes$(\mathcal{A}^{vg}_i,\mathcal{A}^{vg}_j)$; \Comment{\textcolor{blue}{Returns the common nodes between the two vgAMR graphs.}}
        \If {$\mathcal{N}_{\text{common}}$ is empty} \Comment{\textcolor{blue}{The two vgAMRs have no overlapping information.}}
            \State $\mathcal{N}_m = \mathcal{N}_i \cup \mathcal{N}_j \cup \mathcal{N}_\text{multi-sentence}$; \Comment{\textcolor{blue}{Introduce a new, AMR-specific ``multi-sentence" node, to be the root of the merged graph. This node will connect the two disjoint vgAMR graphs.}}
        \Else
            \State $\mathcal{N}'_i = \mathcal{N}_i\setminus \mathcal{N}_{\text{common}}$; $\mathcal{N}'_j = \mathcal{N}_j\setminus \mathcal{N}_{\text{common}}$
            \State $\mathcal{N}_m = \mathcal{N}_{\text{common}} \cup \mathcal{N}'_i \cup \mathcal{N}'_j$
        \EndIf
        \State $\mathcal{E}_m = $ getConnectingEdges$(\mathcal{A}^{vg}_i,\mathcal{A}^{vg}_j,\mathcal{N}_m)$
        \State $\mathcal{A}^{vg}$.remove$(\mathcal{A}^{vg}_i,\mathcal{A}^{vg}_j)$
        \State $\mathcal{A}^{vg}$.add$(\mathcal{A}^{vg}_m)$
    \EndFor
    \State $\mathcal{A}^{vg}_{Meta} = \mathcal{A}^{vg}$ \Comment{\textcolor{blue}{All $N$ vgAMRs are merged into one representation}}
    \State \Return $\mathcal{A}^{vg}_{Meta}$
\end{algorithmic}
\label{alg:metaAMR}
\end{algorithm}

\begin{figure*}[t!]
  \centering
   \includegraphics[width=.95\linewidth]{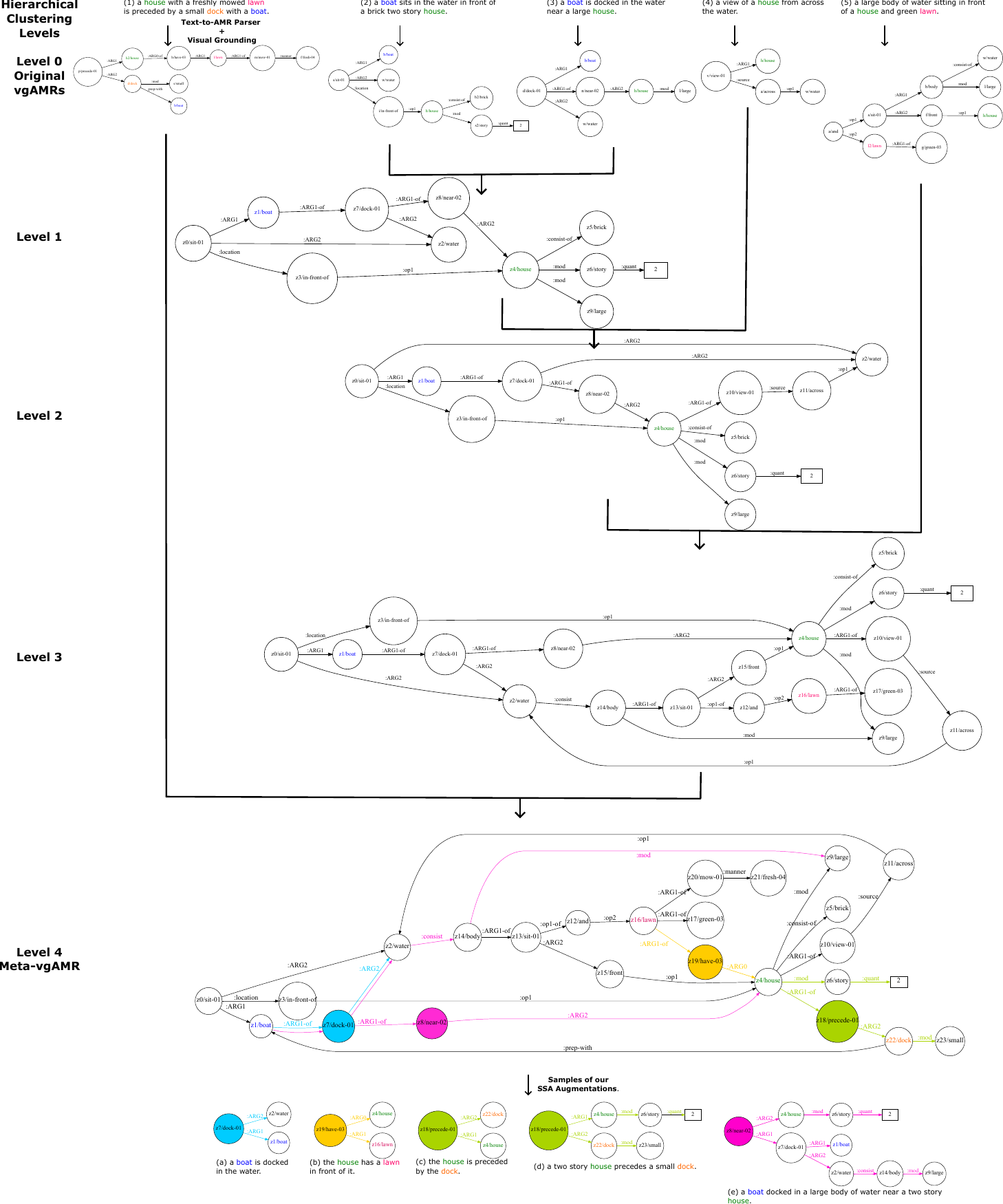}
   \caption{Our Structured Semantic Augmentations process generates new focused captions from visually grounded captions. We derive individual vgAMRs from the original captions and merge them using hierarchical clusters based on the graph similarity of the original vgAMRs (Level 0). At the final layer, we obtain the meta-vgAMR, which combines all available information into one structure. Then, we sample sub-graphs from our meta-vgAMR to generate new captions. Examples (a)-(e) show some of the vgAMRs we sampled with their generated captions.}
   \label{fig:amr-supp}
\end{figure*}

\begin{figure*}[t!]
  \centering
   \includegraphics[width=\linewidth]{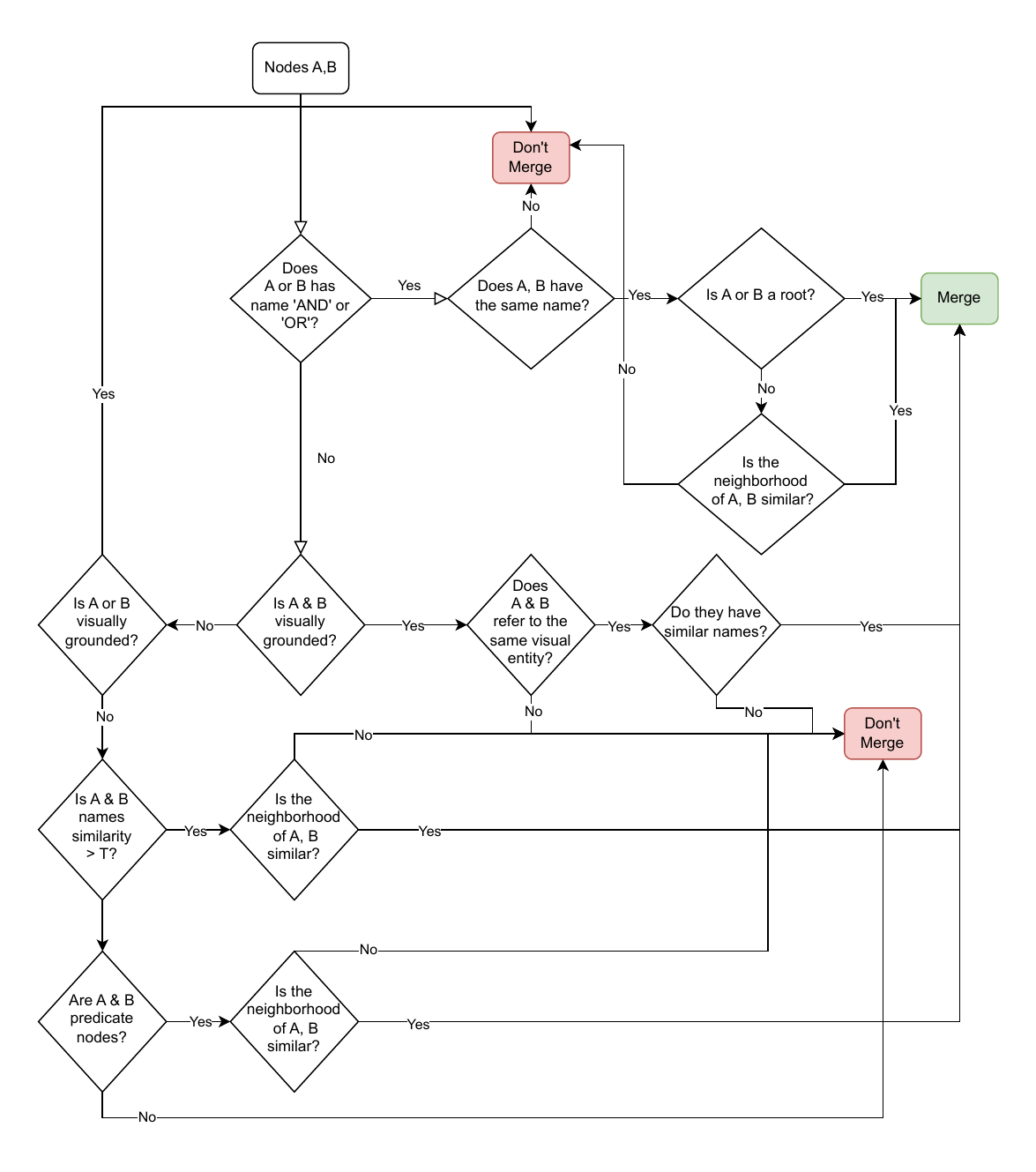}
   \caption{The flow diagram shows the process of merging nodes A and B from two vgAMRs. The merging process starts with handling the AMR-specific nodes of type `AND' and `OR', followed by the visually grounded vgAMR nodes. For the remaining non-grounded nodes (such as predicates, adjectives, and nouns), we check if their names are very similar to decide if they refer to the same concepts and can be merged. If the nodes do not fit the above categories, we check if they are predicate nodes. If they are, we examine their neighborhood to determine if they can be merged.}
   \label{fig:flow-diagram-node-merge}
\end{figure*}

\section{Structured Semantic Augmentation (SSA)}
\label{sec:supp-ssa}
In this section, we will provide additional information on the SSA augmentation strategy we introduced in our main paper. We summarize the steps in constructing the meta-vg Graph (as described in Step 1: Image-level AMR graph generation in our main paper) in \cref{alg:metaAMR}. We include a detailed example of our SSA methodology in \cref{fig:amr-supp}. We present the flow diagram that explains the process of determining if two nodes from different vgAMRs refer to the same concept and, therefore, should be merged in \cref{fig:flow-diagram-node-merge}.

\paragraph{SSA Algorithm.} To construct the hierarchical clusters, we use the UPGMA algorithm, which considers each individual vgAMR as a separate cluster at Level 0. Two clusters are merged at each level based on their distance, starting with the most similar graphs. To measure similarity, we use the Smatch Score between two vgAMR graphs. Since the Smatch score is a metric from 0 to 1, we use 1- Smatch score as the distance metric for the UPGMA algorithm. 
For this example, the AMR graphs of captions (2) and (3) are the most similar, so they are merged first to create their joint vgAMR graph at Level 1. Every graph from levels 1 to 4 results from the 6-17 step of our \cref{alg:metaAMR} where we merge two graphs from lower layers according to the hierarchical clusters computed by UPGMA. The final layer (4) graph is our meta-vgAMR graph, which contains all information from the original vgAMRs and, thus, from the available original captions. 
By applying our event-focused sampling approach, we can generate novel, focused, visually grounded descriptions from this new structure. Some examples can be seen at the bottom of \cref{fig:amr-supp}, along with the resulting captions generated by pre-trained AMR-To-Text parsers.

\paragraph{Finding same-concept nodes in two vgAMR graphs.} In the flow diagram labeled as \cref{fig:flow-diagram-node-merge}, we can observe that when we merge two vgAMR graphs, vgAMR-A and vgAMR-B, we need to identify the common concept nodes between the two representations and combine them. This merging process serves two purposes: a) it allows for a more efficient and compressed representation by reducing redundancies and eliminating multiple nodes for the same concept, and b) it consolidates all available information about a particular concept found in different captions. For example, in \cref{fig:qual-main-supp} c) for the top player, one caption may describe her clothing, another her physical characteristics, and yet another her actions (for instance, practicing martial arts). Despite the differences, all these captions have a common concept: the person in the picture. Instead of having three separate nodes with partial information, we aim to create a single node (person) that consolidates all available information about the person in the image, making it easier to explore all connected nodes and access the complete information.

The process for identifying common nodes involves the steps outlined in \cref{fig:flow-diagram-node-merge}. We start by checking if the two nodes are AMR-specific nodes of type \texttt{AND}. If these nodes are found at the root of the graph, it indicates that the corresponding sentence follows the format `FACT-1 \texttt{AND} FACT-2'. In this case, we can merge them as they represent aggregated facts about the image. If they are not root nodes, we need to be more cautious and ensure that they originate from the same concept. For this reason, we assess the similarity of their neighboring nodes. If they link to the same nodes, we merge them and combine the provided facts. 

As shown in \cref{fig:flow-diagram-node-merge}, if both nodes are visually grounded and refer to the same visual entity (i.e. if they have the same bounding boxes), we are hesitant to merge the nodes without first verifying that their names are synonyms \footnote{We determine if two nodes are synonyms by comparing the cosine similarity of their GloVe embeddings. If the cosine similarity is above a certain threshold, we consider them synonyms.}. This additional condition is helpful in cases where a) the original dataset visually grounds phrases instead of nouns, and b) there is noise from the Text-to-AMR parser. This check ensures that entity attributes (such as `young' and `tall') which may be visually grounded, are not mistakenly merged with noun nodes. Finally, if the two nodes are semantically distant or do not refer to the same visual entity, or if one of them is grounded and the other is not, we conclude that the two nodes cannot be merged.

When we don't have visual cues to help us identify similar concept nodes, we rely on the names of the nodes and their surrounding information to make decisions. If two nodes are synonyms, we look at how similar their neighbors are (e.g., if the two nodes are nouns or adjectives, do they share the same parent?). If they do, we merge them as similar concept nodes.

In our final step, we have an additional procedure for predicate nodes. In our experiments, we observed that the GloVe embeddings of predicate/verb words tend to be more distant. Therefore, in the last step, if the two nodes are predicates, and their child nodes (ARG0, ARG1, and so on) are the same, and the similarity of their names is above a certain threshold (which is smaller than the thresholds used in the previous steps), then we merge the two predicate nodes. This concludes our node merge process.

\section{Experimental Setup}
\label{sec:exp-setup-supp}

\subsection{Evaluation Metrics}
\label{sec:metrics-supp}

\paragraph{\bf Content Controllability: IoU.} To measure {\it content controllability} we design an extended version of the IoU metric of \cite{cornia2019show} that calculates the degree-of-match (faithfulness) between a control signal and the corresponding generated caption. For our control signal, we use the set of nouns $\mathcal{E}$ that represent the entities of interest, which are the names of the visual objects in the control. To extract nouns from the predicted sentences, we use the Stanford part-of-speech tagger \cite{toutanvoa2000enriching, toutanova2003feature}. We then find the semantic intersection of the two sets using Hungarian Matching, as in \cite{cornia2019show}. Finally, we calculate the semantic intersection over union of the control nouns and the nouns extracted from the controllable caption, which gives us our content controllability IoU. 

In particular, we measure content controllability using the IoU (the overlap between two sets) of the set of nouns in the control signal and the set of nouns in the generated sentence.
\begin{itemize}
    \item For the set of control signal nouns $\mathcal{E}$ in \cocoent\ test set, we use the existing annotations. For each caption, the head noun of a noun chunk is provided. We use the set of head nouns as our $\mathcal{E}$.
    \item For \flickrent\ this information is not available. We use the object labels from Faster R-CNN to get the control signal nouns.
\end{itemize}

Our IoU metric is based on the corresponding score in \cite{cornia2019show}. We modify the following parts:
\begin{itemize}
    \item Ground truth nouns: Instead of using the ground truth captions as a proxy, we directly extract control signal nouns from the control signal itself, as described in the previous bullet points.
    \item Generated sentence nouns: For the generated controlled sentence, instead of looking if each word is in a dictionary of nouns prepared by \cite{cornia2019show}, we use part-of-speech tagging \cite{toutanvoa2000enriching, toutanova2003feature} to extract the nouns of the sentence. We use this approach because the provided dictionary, although it contained many nouns, was not a complete list, so in many cases, during evaluation, nouns were discarded because there was no entry for them in the dictionary, which added noise to the original metric.
\end{itemize}
Our next steps are as described in \cite{cornia2019show}, that is, the Hungarian matching of the two sets of nouns using the cosine similarity of the corresponding GLoVE embeddings for each noun word. The final IoU is the sum of cosine similarities for the aligned nouns. 

The advantage of our IoU score from the one proposed in \cite{cornia2019show} is that it directly compares the control signal with the generated sentence; instead of the dataset ground truth sentences, which are just a proxy of entities in the control signal. This helps reduce the metric noise, since our IoU are not affected from annotation errors (for example, ground truth captions where not all entities are annotated/grounded to a bounding box, which will lead to a noisy proxy of the control signal) or from missing entries in the noun dictionary used in \cite{cornia2019show}.

\paragraph{\bf Content Controllability: Hallucinations. } We propose the Hallucinating Nouns (Hal) content controllability metric to help us to determine the number of hallucinations present in the generated captions. These hallucinations refer to nouns or visual entities that are not part of the control signal. They could be visual objects present in the image but not in focus of the control signal, or visual objects that are not present in the image at all. To measure this, we propose the `Hal-lucinating Nouns' metric, which can be computed using the following equation:
\begin{align}
    \mathrm{Hal} &= \frac{1}{|\mathcal{N}|}
    \left( |\mathcal{N}| - \text{IoU}(\mathcal{N},\mathcal{E}) \right).
\end{align}
where $\mathcal{N}$ is the set of nouns extracted from the generated controlled caption and $\mathcal{E}$ is the set of nouns (visual entities) in the control signal. 

\paragraph{\bf Diversity.} To measure {\it diversity}, we compute n-gram diversity, D-$n$ for $n=1,\,2$ \cite{aneja2019sequential}, as well as self-CIDEr-based diversity (sC) \cite{wang2019describing}. 
D-$n$ measures the ratio of distinct $n$-grams to the total number of words generated per set of diverse captions. sC computes the diversity of a set of captions by using their CIDEr score \cite{vedantam2015cider}, a metric that measures sentence similarity by giving more weight to the matching of novel words. 
For a fair comparison of the different CIC models, 
we measure diversity for the five generated captions for each test image (in \cocoent and \flickrent), and report their average. Note that not all images in \cocoent and \flickrent have five caption--control signal pairs, especially for \cocoent that is automatically annotated. We only considered the ones with five available pairs for diversity evaluation, including $985$ images for \flickrent and $112$ images for \cocoent.

\paragraph{\bf Best-5 Diversity.} For completeness, we compute the best-5 diversity, proposed in \cite{chen2020say}. Specifically, we generate $M=10$ randomly generated control signals for a given image. From the $M$ captions, we form all possible sets of $5$ captions ($M$ choose $5$) and measure the ratio of $n$-grams to the total number of words for each set. We report the average of the best Div-$n$ scores for all images in the test set.  

\paragraph{\bf Length Controllability. } For {\it length controllability} (L), we measure the Mean Absolute Error (MAE) between the fine length control (number of words) and the size of the resulting $M=10$ controlled captions, which are generated from $M$ randomly created control signals. We also calculate the length precision (LP) \cite{deng2020length} by determining the percentage of generated captions that match the desired coarse length level.  

\paragraph{\bf Text Quality. } We assess {\it text quality} of generated captions using GRUEN (G) \cite{zhu2020gruen}, a reference-free metric based on BERT contextual embeddings that measure the syntactic and semantic well-formedness of a text segment.

\paragraph{\bf Overall Performance using Harmonic Means.} Finally, we measure the {\it overall performance} of each model based on its ability to balance content controllability, diversity, and text quality. To calculate this, we use the harmonic mean of IoU, G, and sC. All of these metrics range between 0 and 1, with a higher value indicating better performance. The harmonic mean ($H$) helps us determine the model with the best overall performance. It prioritizes models that perform well across all metrics while penalizing those with poor performance, even in one metric.

\paragraph{\bf Standard Captioning Metrics.} Following prior work, we also report performance with respect to standard captioning metrics, namely Bleu-4 (B4) \cite{papineni2002bleu}, Meteor (M) \cite{banerjee2005meteor}, Rouge (R) \cite{lin2004rouge}, CIDEr (C) \cite{vedantam2015cider} and Spice (S) \cite{anderson2016spice}. Specifically, Bleu measures the n-gram similarity of the two sentences, but without examining their synonymity, something that is addressed by Meteor. Rouge estimates the recall of their largest common subphrase. CIDEr score gives more weight to the matching of novel words and finally, Spice computes the semantic similarity by comparing the scene graph representations of the two sentences. For all these measures, higher is better. 
Note that these metrics are not sufficient for evaluating CIC, as they compare a generated (controlled) caption with a reference ground-truth caption, ignoring the desired effect of the control signal. We report them for completeness, but we believe that our well-formedness metric (GRUEN) is more suited for evaluating the quality of the controlled captions.

\subsection{CIC-BART-SSA and Baselines Setup}
\label{sec:sup-setup-ssa-baseline}
\paragraph{\bf SSA Parameters.} For our AMR-to-Text generated sentence filtering, we set the GRUEN (G) threshold to 0.7.

\paragraph{\bf Model Parameters.} We initialize \modelname encoder and decoder from the pre-trained weights of VL-BART \cite{cho2021unifying} to benefit from transfer learning. 
We further train our model on data that contains image--control--caption triplets, where control consists of the above-mentioned signals. 
We incorporate five different length levels to control the length of our output. Each level has a specific range, with level A ranging from one to nine words, level B spanning ten to nineteen words, level C covering twenty to twenty-nine words, level D consisting of thirty to thirty-nine words, and level E including sentences with forty or more words. In our \modelname vocabulary, we have added five tokens to represent these five caption length levels. 
For optimizing the cross-entropy loss, we utilize the RAdam optimizer \cite{liu2019variance} with a learning rate of $5\cdot10^{-5}$ and a batch size of 80. We train our models for 20 epochs and select our trained model based on the best content controllability IoU and CIDEr scores.

\begin{figure*}[ht]
  \centering
    \includegraphics[width=.95\linewidth]{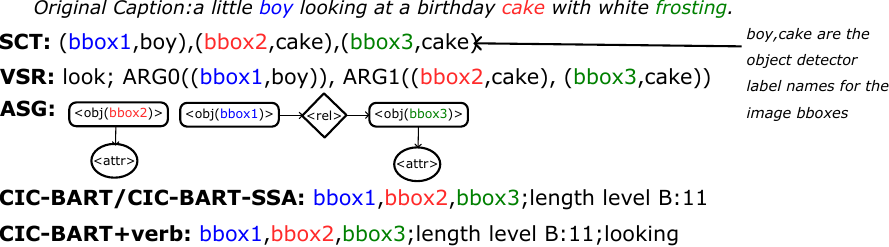}
    \caption{An illustrative example of the used control signals from different CIC models. 
    }
    \label{fig:types-of-control}
\end{figure*}

\paragraph{\bf Baseline Models.} 
We conducted evaluations for metrics such as content controllability (IoU), text quality (G), and diversity (D-1, D-2 and sC) for SCT and VSR, using the code and pre-trained checkpoints available on their official project GitHub pages. However, for ASG, we re-trained and evaluated the ASG2Caption model for \cocoent using the official GitHub codebase since the pre-trained checkpoints were not available. Unfortunately, we could not train the ASG2Caption model on the \flickrent dataset as the ASG dataset for \flickrent has not been released. For the standard captioning metrics, best-5 diversity, and length precision of the testing sets of the datasets \cocoent\ and \flickrent, we used the values presented in the corresponding papers. 

We mention, that in our main paper, we used the strongest model performance from ComPro, which employs GPT-2 Large. This model has a total of 881M parameters, 107M of which are used for the mapping network and 774M are from GPT-2. It's worth noting that our models, namely, \modelname, \modelname-SSA, and \modelname+verb, use only 140M parameters, making them more than six times smaller than ComPro with GPT-2 Large.

In \cref{fig:types-of-control}, we present an example with the control signals used by the baselines and our models for a specific instance where we need a focused caption on the boy (bbox 1) and the cake (bbox2, bbox3). The SCT model \cite{cornia2019show} uses bounding boxes of entities of interest and GLoVE embeddings of their Faster R-CNN labels as control signals. The VSR model \cite{chen2021human} adds ground truth caption verbs and their PropBank grounded verb semantic roles to the SCT control signal. The ASG model \cite{chen2020say} employs abstract scene graphs as control signals that provide information about how visual entities are related or connected and how many attributes they have. 
Our models (\modelname and \modelname-SSA) use only the bounding boxes of interest and the desired caption length level as control signals. We have also explored the use of ground truth verbs in the control signal, like in VSR, in our \modelname+verb model. However, unlike VSR, we only use the ground truth verb name and not their PropBank grounded semantic roles.

\section{Results}
\label{sec:result-supp}

\begin{figure*}[ht]
  \centering
   \includegraphics[width=\linewidth]{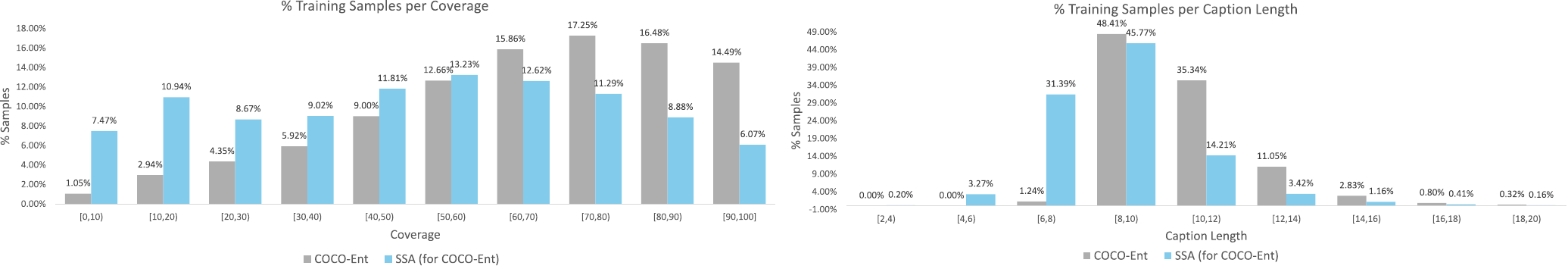}
   \caption{Training set coverage and caption length histograms of \cocoent and SSA augmentations for this dataset. Coverage represents the \% of the image covered by the control signal, so the left and right parts of the graph represent more focused and broader control signals, respectively. We note that the \cocoentssa dataset will contain both the original data (gray bars) and our SSA (blue bars).}
   \label{fig:coco-train-histograms}
\end{figure*}
\begin{figure*}[ht]
  \centering
   \includegraphics[width=\linewidth]{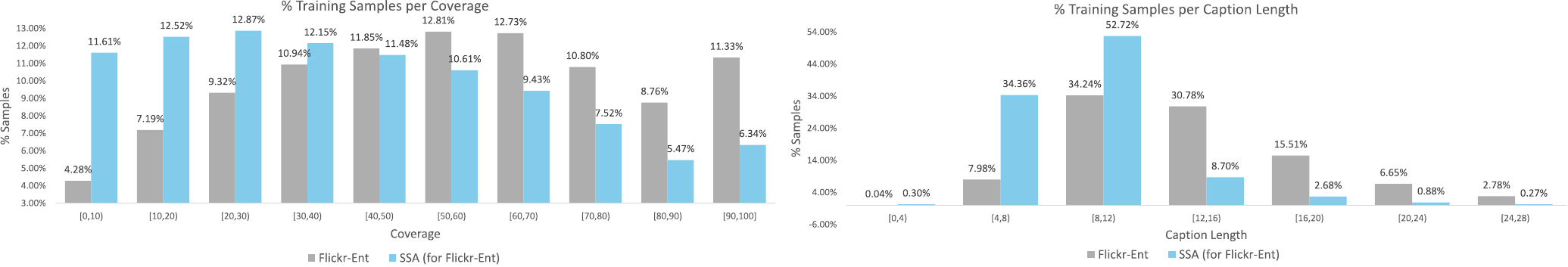}
   \caption{Training set coverage and caption length histograms of \flickrent and SSA augmentations for this dataset. Coverage represents the \% of the image covered by the control signal, so the left and right parts of the graph represent more focused and broader control signals, respectively. We mention that \flickrssa dataset contains both the original (gray bars) and SSA (blue bars) samples.}
   \label{fig:flickr-train-histograms}
\end{figure*}

\subsection{Original and SSA Augmented Datasets Analysis}\label{sec:datastats}
In \cref{fig:coco-train-histograms,fig:flickr-train-histograms}, we present the coverage and caption length statistics of the \cocoent and \flickrent training set and their derived SSA augmentations, respectively.
When analyzing the scene coverage based on the control signal, it becomes apparent that the original datasets predominantly feature samples that describe the entire image (high coverage), with very few focusing on a small portion of the scene (highly focused control signals, low coverage). This is particularly evident in the \cocoent dataset, where examples with focused control signals are minuscule. 
For the caption length statistics, we notice that \cocoent dataset is far from diverse with approximately 84\% of the captions having 8-12 words. Similarly, in the \flickrent dataset, approximately 65\% of its descriptions have 8-16 words.

With our SSA data (blue bars), we augment the original datasets (gray bars) with highly focused control--caption pairs and diverse caption length, to construct a new dataset of spatially and linguistically diverse data for controllable image captioning, namely the \cocoentssa and \flickrssa datasets.

\subsection{Original and SSA captions Mixtures}\label{sec:supp-ssamix-strategy}
In this section, we delve deeper into the impact of our SSA augmentation on CIC models. To conduct our experiments, we utilize our mixing methodology described in our main paper (Section 4.1), to create various versions of \cocoentssa and \flickrssa datasets. We aim to examine the effect of our SSA examples, so we include all original samples in the mixed dataset $\mathcal{D}_{SSA}$. Formally, we state that $\sam_{\mathcal{D}}(\tau_{\mathcal{D}}, p_{\mathcal{D}}) = \mathcal{D}$.

We conducted six experiments for our augmentation function $\sam_{SSA}$. In the first scenario, we randomly sampled $x\%$ of the SSA samples. This is equivalent to the `Random Sampling Strategy' (R). To test the impact of parameter $p_{SSA}$, we experimented with the following percentages: 0\% (no SSA samples); 25\% (all original and a random 25\% of the generated SSA samples); 50\%; 75\%; and 100\% (all original and all generated SSA samples). Additionally, we conducted an experiment using the `Uniform Coverage Sampling Strategy' (U) for $\tau_{SSA}$, with $p_{SSA}$ set to ten (10) uniform coverage bins ($p_{SSA} = \{ [0\%,10\%), [10\%,20\%), \dotsc, [90\%,100\%] \}$). We note that the `Uniform Coverage Sampling Strategy' contains approximately the same number of SSA samples as the 50\% random sampling strategy. 

\begin{table}[ht]
\caption{
Content (IoU, Hal) and length (L) controllability, text quality (G), diversity (D-1, D-2, sC), and harmonic mean (H) of (IoU, G, and sC) for our \modelname-SSA models evaluated only on the original \flickrent test set. Each of our \modelname-SSA models is trained on a different \flickrssa mixture, described by the augmentation strategy type $\tau_{SSA}$ and parameters $p_{SSA}$. Each blended data version has all the original data but different percentages of our SSA augmentations. The row order of experiments corresponds to the included SSA percentage in \flickrent ranging from 0 to 100\%.
}
\scriptsize
\centering
\setlength{\tabcolsep}{4pt}
\begin{tabular}{@{}cc|>{\columncolor[gray]{.9}}cccccccc}
\toprule
\multicolumn{10}{c}{Model: \modelname-SSA | Evaluation: \flickrent Test Set} \\
\midrule
$\tau_{SSA}$ & $p_{SSA}$ & $H\uparrow$ & IoU$\uparrow$ & G$\uparrow$ & sC$\uparrow$ & L$\downarrow$ & Hal$\downarrow$ & D-1$\uparrow$ & D-2$\uparrow$ \\
\midrule
R & 0\%  & 69.8 & 54.0 & 85.0 & 78.6 & 1.24 & 36.5 & 43.6 & 58.2\\
R & 25\% & 69.9 & 53.7 & 85.1 & 79.8 & 1.29 & 36.5 & 44.7 & 59.5\\
R & 50\% & 70.3 & 53.9 & 85.6 & 80.5 & 1.23 & 36.2 & 45.3 & 60.6\\
U & 10   & 70.6 & 54.3 & 85.6 & 80.5 & 1.07 & 35.6 & 45.9 & 61.0  \\
R & 75\% & 70.5 & 53.9 & 85.5 & 81.1 & {\bf1.05} & 35.9 & 46.2 & 61.7\\
R & 100\%& \bf{71.3} & \bf{55.0} & \bf{86.0} & \bf{81.7} & {\bf1.05} & {\bf34.1} & {\bf47.0} & {\bf62.6}\\
\bottomrule
\end{tabular}
\label{tab:flickr-mix-str-supp}
\end{table}

\begin{table}[ht]
\caption{
Content (IoU, Hal) controllability, text quality (G), diversity (D-1, D-2, sC), and harmonic mean (H) of (IoU, G, and sC) for our \modelname-SSA models evaluated only on the SSA data generated using \flickrent test set. Each of our \modelname-SSA models is trained on a different \flickrssa mixture, described by the augmentation strategy type $\tau_{SSA}$ and parameters $p_{SSA}$. Each blended data version has all the original data but different percentages of our SSA augmentations.  The row order of experiments corresponds to the included SSA percentage in \flickrssa ranging from 0 to 100\%.
}
\scriptsize
\centering
\setlength{\tabcolsep}{4pt}
\begin{tabular}{@{}cc|>{\columncolor[gray]{.9}}ccccccc}
\toprule
\multicolumn{9}{c}{Model: \modelname-SSA | Evaluation: SSA Test Set} \\
\midrule
$\tau_{SSA}$ & $p_{SSA}$ & $H\uparrow$ & IoU$\uparrow$ & G$\uparrow$ & sC$\uparrow$ & Hal$\downarrow$ & D-1$\uparrow$ & D-2$\uparrow$    \\
\midrule
 R & 0\%   & 68.5 & 53.0 & 80.5 & 79.8  & 37.3 & 52.9 & 62.9 \\
 R & 25\%  & 70.5 & 54.4 & 81.4 & 84.0  & 35.3 & 55.5 & 67.2 \\
 U & 10    & 71.3 & 55.4 & 82.6 & 83.7  & 33.9 & 56.1 & 67.4 \\
 R & 50\%  & 71.5 & 55.2 & 82.7 & 85.1 & 33.9 & 56.4 & 68.3 \\
 R & 75\%  & 71.6 & 55.4 & 82.8 & 84.8 & 33.3 & 56.4 & 68.7 \\
 R & 100\% & {\bf72.0} & {\bf55.6} & {\bf82.9} & {\bf86.1} & {\bf33.0} & {\bf56.5} & {\bf69.3} \\
\bottomrule
\end{tabular}
\label{tab:ssa-flickr-mix-str-supp}
\end{table}

\begin{table}[ht]
\caption{Content (IoU, Hal) and length (L) controllability, text quality (G), diversity (D-1, D-2, sC) and harmonic mean (H) of (IoU, G, and sC) for our \modelname-SSA models evaluated only on the original \cocoent test set. Each of our \modelname-SSA models is trained on a different \cocoentssa mixture, described by the augmentation strategy type $\tau_{SSA}$ and parameters $p_{SSA}$. Each blended data version has all the original data but different percentages of our SSA augmentations. The row order of experiments corresponds to the included SSA percentage in \cocoentssa ranging from 0 to 100\%.
}
\scriptsize
\centering
\setlength{\tabcolsep}{4pt}
\begin{tabular}{@{}cc|>{\columncolor[gray]{.9}}cccccccc}
\toprule
\multicolumn{10}{c}{Model: \modelname-SSA | Evaluation: \cocoent Test Set} \\
\midrule
 $\tau_{SSA}$ & $p_{SSA}$ & $H\uparrow$ & IoU$\uparrow$ & G$\uparrow$ & sC$\uparrow$ & L$\downarrow$ & Hal$\downarrow$ & D-1$\uparrow$ & D-2$\uparrow$ \\
\midrule                                             
 R & 0\%  & 75.9      & 76.2      & 73.0      & 78.7      & .490      & 19.0      & 38.0      & 56.2 \\
 R & 25\% & 76.9      & 76.5      & 74.6      & 80.1      & .148      & 18.5      & 42.6      & 59.7 \\
 R & 50\% & 76.8      & 76.7      & 74.9      & 79.0      & .163      & \bf{17.8} & 42.9      & 59.0 \\
 U & 10   & 76.7      & 77.0      & 74.0      & 79.4      & .150      & {\bf17.8} & 42.0      & 58.6 \\
 R & 75\% & 77.8      & \bf{77.2} & 74.0      & {\bf82.6} & .116      & {\bf17.8} & 42.9      & 61.6\\
 R & 100\%& \bf{78.3} & \bf{77.2} & \bf{74.8} & 82.5      & {\bf.106} & {\bf17.8} & \bf{44.6} & \bf{63.2}\\
\bottomrule
\end{tabular}
\label{tab:coco-mix-str-supp}
\end{table}

\begin{table}[ht]
\caption{
Content (IoU, Hal) controllability, text quality (G), diversity (D-1, D-2, sC), and harmonic mean (H) of (IoU, G, and sC) for our \modelname-SSA models evaluated only on the SSA data generated using \cocoent test set. Each of our \modelname-SSA models is trained on a different \cocoentssa mixture, described by the augmentation strategy type $\tau_{SSA}$ and parameters $p_{SSA}$. Each blended data version has all the original data but different percentages of our SSA augmentations. The row order of experiments corresponds to the included SSA percentage in \cocoentssa ranging from 0 to 100\%.
}
\scriptsize
\centering
\setlength{\tabcolsep}{4pt}
\begin{tabular}{@{}cc|>{\columncolor[gray]{.9}}ccccccc}
\toprule
\multicolumn{9}{c}{Model: \modelname-SSA | Evaluation: SSA Test Set} \\
\midrule
 $\tau_{SSA}$ & $p_{SSA}$ & $H\uparrow$ & IoU$\uparrow$ & G$\uparrow$ & sC$\uparrow$ & Hal$\downarrow$ & D-1$\uparrow$ & D-2$\uparrow$    \\
\midrule
 R & 0\%  & 69.2      & 61.4      & 73.9      & 74.0      & 28.4      & 44.2 & 57.0\\
 R & 25\% & 74.4      & 65.1      & 80.1      & 80.0      & \bf{23.0} & 50.1 & 63.1\\
 U & 10   & 74.6      & 65.0      & 80.3      & 80.8      & 23.2      & 51.2 & 64.4\\
 R & 50\% & 74.9      & 64.9      & {\bf80.7}      & 81.7      & 23.1      & 51.7 & 64.3\\
 R & 75\% & 74.9      & 64.9      & {\bf80.7} & 81.6      & 23.1      & 51.7 & 65.1\\
 R & 100\%& {\bf75.6} & {\bf65.2}      & {\bf80.7}      & {\bf83.7} & 23.2      & {\bf53.8} & {\bf67.8}\\
\bottomrule
\end{tabular}
\label{tab:ssa-coco-mix-str-supp}
\end{table}

We repeat the procedure for both the \cocoentssa and \flickrssa datasets. We evaluate all models on content (IoU, Hal) and length (L) controllability, text quality (G), diversity (sC, D-1, D-2), and the harmonic mean (H) of IoU, G, and sC. 
We present the evaluation results for the \cocoent and \flickrent test sets in \cref{tab:coco-mix-str-supp,tab:flickr-mix-str-supp}. We observe a similar trend in both datasets where adding our SSA samples improves context and length controllability, text quality, and diversity. Our significant improvement in diversity and length controllability is due to the linguistic diversity offered by our SSA augmentations. For example, in \cref{fig:coco-train-histograms} caption length histogram, we can see how narrow it is for \cocoent dataset which mainly contains captions with 11, 12, or 13 words, so it is difficult for models trained on just \cocoent to generalize and generate captions of other lengths. On the contrary, our models trained jointly with our SSA augmentations can generate captions faithful to the length control signal.

In \cref{tab:ssa-coco-mix-str-supp,tab:ssa-flickr-mix-str-supp}, we have evaluated the performance of each model on the SSA augmentations from the \cocoent and \flickrent testing images, respectively. We have excluded the length (L) controllability as it is the same as in \cref{tab:coco-mix-str-supp,tab:flickr-mix-str-supp}. This is because it is computed on random control signals of each dataset testing images. 

We observe an even more evident improvement across all metrics as we progressively include more of our focused SSA examples for CIC model training. This results from our SSA augmentations, which provide focused examples for training controllable image captioning models. This is exemplified in \cref{fig:coco-train-histograms,fig:flickr-train-histograms} coverage histograms, where the low coverage (high focus) regime is highly under-represented in the original \cocoent and \flickrent datasets.
Finally, our quantitative analysis demonstrates that training with our SSA augmentations improves controllability, text quality, and diversity performance. Particularly, the improvement is significant in cases where the CIC models need to focus on and describe a specific, small region of a complex and large scene.

\subsection{Effect of SSA on Content Controllability}\label{sec:ssaciceffects}
\begin{figure*}[ht]
    \centering
    \includegraphics[width=\linewidth]{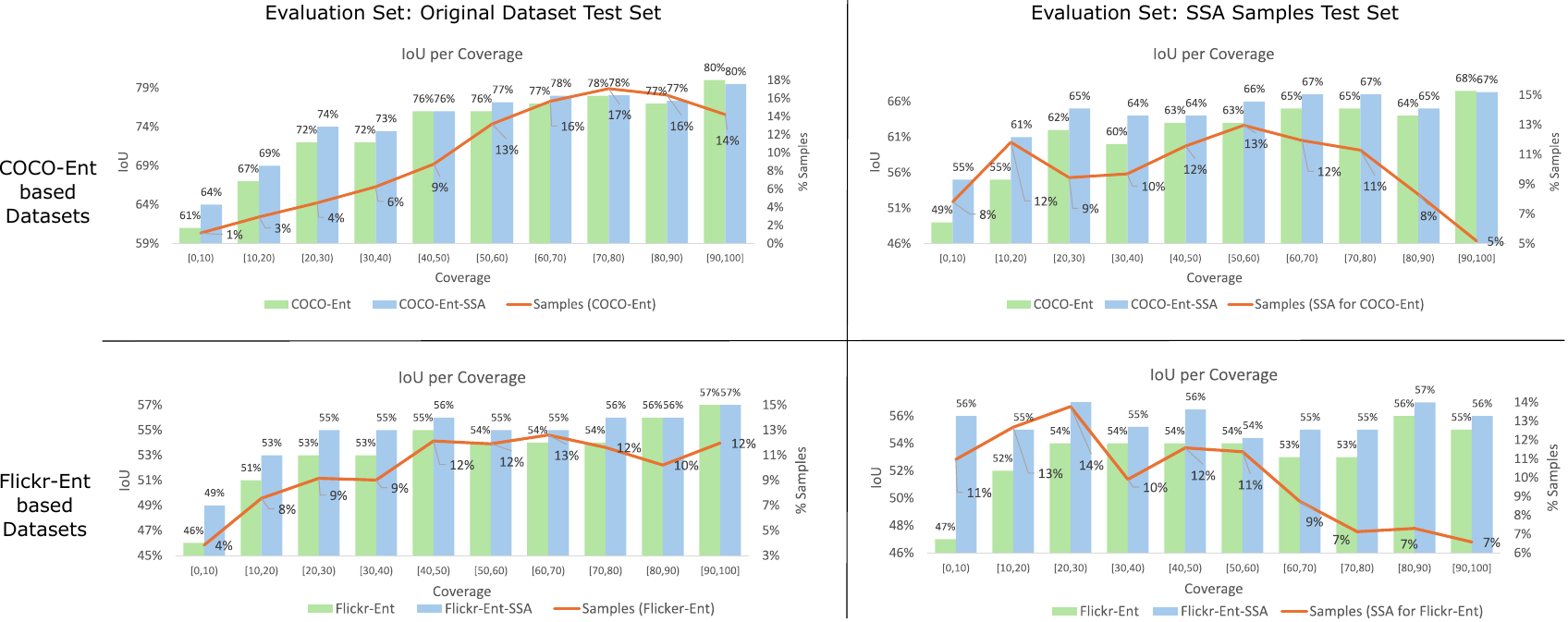}
    \caption{Content controllability (IoU) performance of \modelname when trained with and without SSA. The abscissa is the \% of the image covered by the control signal, so the left and right parts of the graph represent more focused and broader control signals, respectively. The \%Samples curve represents the distribution of \flickrent test images in each coverage interval. The results show that SSA plays a crucial role in boosting \modelname performance in data-deprived, focused CIC settings.
    }
    \label{fig:iou-coverage-analysis-all-supp}
\end{figure*}
\begin{figure*}[ht]
    \centering
    \includegraphics[width=\linewidth]{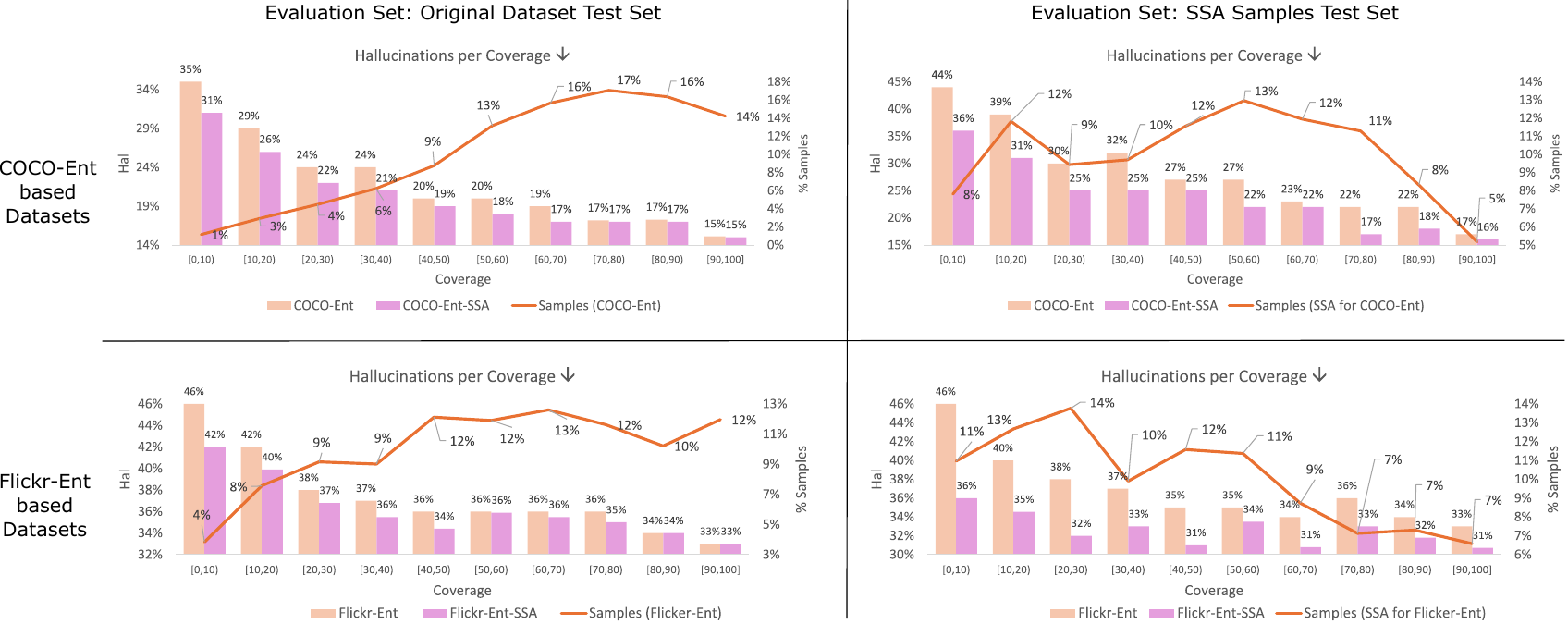}
    \caption{Hallucinating Nouns (Hal) performance of \modelname when trained with and without SSA. The abscissa is the \% of the image covered by the control signal, so the left and right parts of the graph represent more focused and broader control signals, respectively. The \%Samples curve represents the distribution of \flickrent test images in each coverage interval. The results show that SSA plays a crucial role in boosting \modelname performance in data-deprived, focused CIC settings.
    }
    \label{fig:hal-coverage-analysis-all-supp}
\end{figure*}
In this section, we present the performance of our models, namely \modelname and \modelname-SSA, with regards to IoU (Intersection over Union) analysis. The former is trained on the original datasets, \cocoent and \flickrent, while the latter is trained on our proposed datasets, \cocoentssa and \flickrssa. We break down the content controllability (IoU, Hal) performance on coverage bands, where coverage refers to the percentage of the image covered by the control signal. 

In \cref{fig:iou-coverage-analysis-all-supp}, we show the results of our models on different evaluation test sets. The first row of the figure represents the performance of models trained on either \cocoent or \cocoentssa datasets, while the second row represents models trained on \flickrent or \flickrssa. The two columns describe the evaluation test sets. In the first column, we evaluate the models on the original datasets (\cocoent, \flickrent) test sets. In contrast, in the second column, we evaluate our SSA augmentations derived from the test sets of the original datasets. 

The orange line in all plots represents the percentage of examples in each coverage band. We observe that the test sets of the original datasets, \cocoent and \flickrent, have more examples with control signals covering a broad aspect of the image. In contrast, the SSA test sets have more samples for focused control signals covering a small percentage of the image. We also notice that the test sets of the original and SSA datasets are consistent with their respective training set statistics presented in \cref{fig:coco-train-histograms,fig:flickr-train-histograms}.

Furthermore, in \cref{fig:hal-coverage-analysis-all-supp}, we present the corresponding coverage histograms for our Hallucinations (Hal) metric. We observe a similar trend in all cases, wherein our model \modelname-SSA, trained on our SSA augmentations, shows an improvement in content controllability performance. This means higher IoU and reduced Hal. We note that breaking down the content controllability metrics in coverage bands reveals the major improvement in the low coverage regions, where the original datasets, \cocoent and \flickrent, have very few data points.

\begin{table}[ht]
\caption{Best-5 Diversity for randomly generated control signals of the \cocoent and \flickrent testing images. Our model \modelname was trained on the original \cocoent and \flickrent datasets while \modelname-SSA was trained with our \cocoentssa and \flickrssa augmented datasets. \textit{*ASG-type dataset was not released for \flickrent, precluding us from evaluating its best-5 diversity scores.}}
\scriptsize
\centering
\setlength{\tabcolsep}{4pt}
\begin{tabular}{@{}lcc|cc}
\toprule
Method & D-1$\uparrow$ & D-2$\uparrow$ & D-1$\uparrow$ & D-2$\uparrow$ \\
\midrule
 &\multicolumn{2}{c}{\cocoent} & \multicolumn{2}{c}{\flickrent}\\
ASG \cite{chen2020say} & 43 & 56 & - & -\\
\modelname & 58 & 86 & 67 & 90\\
\modelname-SSA & {\bf67} & {\bf92} & {\bf68} & {\bf93}\\
\bottomrule
\end{tabular}
\label{tab:best-5-div-supp}
\end{table}

\subsection{Best-5 Diversity} \label{sec:diversity}

In \cref{tab:best-5-div-supp}, we present the best-5 D-1, D-2 diversity for our models \modelname and \modelname-SSA, which was proposed in ASG \cite{chen2020say}. We notice an important diversity improvement, especially for the \cocoent dataset, when we train our models using our SSA augmentations (\modelname-SSA).

\subsection{Measuring Performance via Standard Captioning Metrics} \label{sec:standard-captioning}

\begin{table}[ht]
\caption{Captioning metrics on \cocoent and \flickrent\ original test sets.
}
\centering
\scriptsize
\setlength{\tabcolsep}{2.5pt}
\begin{tabular}{@{}lccccc|ccccc}
\toprule
Model        & B4$\uparrow$   & M$\uparrow$   & R$\uparrow$   & C$\uparrow$   & S$\uparrow$ & B4$\uparrow $  & M$\uparrow$   & R$\uparrow$   & C$\uparrow$   & S$\uparrow$  \\
\midrule
&\multicolumn{5}{c}{\cocoent} & \multicolumn{5}{c}{\flickrent}  \\
LaBERT\cite{deng2020length}     & 13.5 & 20.6 & 42.3 & 136.6 & 32.4 & 8.1 & 14.6 & 32.7 & 70.8 & 19.6\\
SCT\cite{cornia2019show}        & 22.3      & 25.6      & 55.3      & 209.7      & 48.5        & 12.5      & 16.8      & 38.9      & 84.0      & 23.5       \\
ComPro\cite{wang2023learning}     & 24.0 & 27.3      & 56.1      & 232.2   & \underline{50.4}        & 11.9      & 17.3      & 37.8      & 89.4      & 23.9       \\
ASG\cite{chen2020say}             & 23.0    & 24.5     & 50.1     & 204.2      & 42.1        & - & - & - & - & -  \\
VSR\cite{chen2021human}           & \underline{25.4} & \underline{28.8}      & \underline{57.8}      & \underline{265.0}      & 49.8       & 12.3      & \underline{19.8}      & \underline{40.9}      & 131.4      & 22.4       \\
\modelname                        & 21.0      & 26.2      & 50.2      & 225.0     & 46.3        & \underline{14.2}      & 19.4      & 39.7      & \underline{136.4}      & \underline{27.2}       \\
\modelname-SSA                    & 20.0      & 25.5      & 48.9      & 216.2      & 46.1        & 13.0      & 18.9     & 37.8      & 123.3      & 27.0       \\
\modelname+verb                    & \bf{36.2} & \bf{33.7} & \bf{62.9} & \bf{366.8} & \bf{53.7}   & \bf{26.6} & \bf{27.2} & \bf{53.9} & \bf{275.1} & \bf{32.4}  \\
\bottomrule
\end{tabular}
\label{tab:captioning}
\end{table}

\cref{tab:captioning} reports the results of all models in standard captioning metrics (i.e., B4, M, R, C and S). As we can see, both \modelname and \modelname-SSA perform comparably to the three baselines with respect to these metrics. Nonetheless, as we noted earlier, these scores reflect the match between a generated controlled caption and a ground-truth image-level caption. Given a focused control signal (e.g., one focusing on a subset of entities in an image), we expect a partial match between the generated controlled caption and the ground-truth caption. VSR has the best scores for most of these metrics, but this is partially due to this model using the exact verb as the control signal and is not necessarily an indicator of this model's caption quality (as we saw earlier with the low G score). To understand the role of such descriptive control signals, we present results for a variation of our \modelname where we also input the verb as an additional control signal; see the last row of the table (\modelname+verb). Note that even with this additional information, our control signal is still simpler than that of the VSR, as we do not provide the verb-specific semantic roles. Nevertheless, by adding a verb as the control signal, we can see a substantial increase in all standard captioning metrics.

\begin{figure*}[ht]
  \centering
    \includegraphics[width=\linewidth]{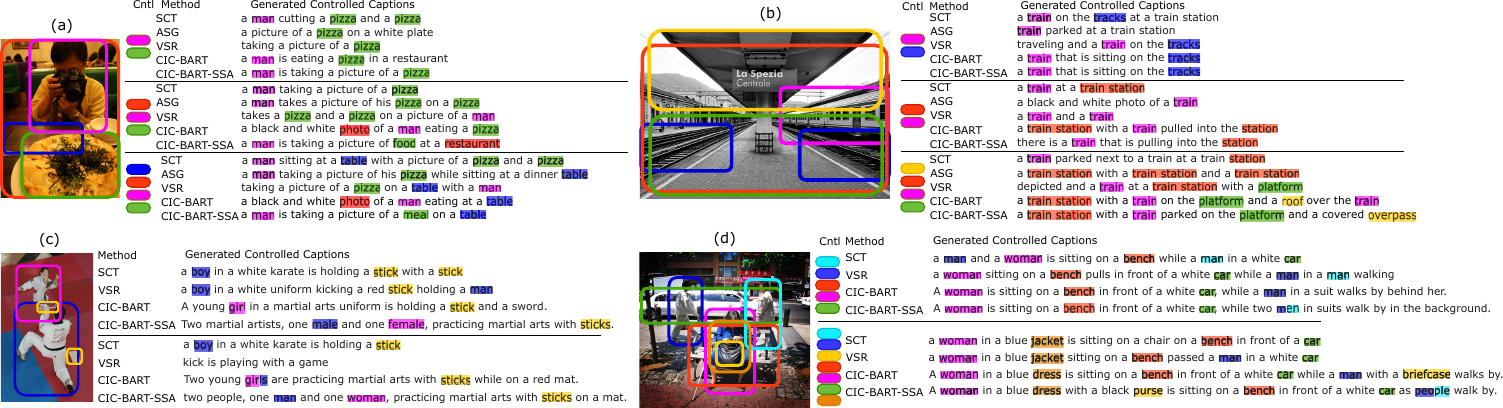}
  \caption{Qualitative examples of generated controllable captions from \cocoent (images (a) and (b)) and \flickrent (images (c) and (d)) test set. 
  Images (a) \url{https://farm5.staticflickr.com/4102/4888234256\_538b8dee56\_z.jpg} and (b) \url{https://farm9.staticflickr.com/8501/8308004994\_44eb2d562d\_z.jpg} are licensed under a Creative Commons CC BY-SA 2.0 \url{https://creativecommons.org/licenses/by-sa/2.0/}; (c) \url{flickr.com/photo.gne?id=101362133} and (d) \url{flickr.com/photo.gne?id=151970521} are licensed under a Creative Commons CC BY 2.0 \url{https://creativecommons.org/licenses/by/2.0/}.
  }
  \label{fig:qual-main-supp}
\end{figure*}

\begin{figure}[t!]
    \centering
    \includegraphics[width=.8\linewidth]{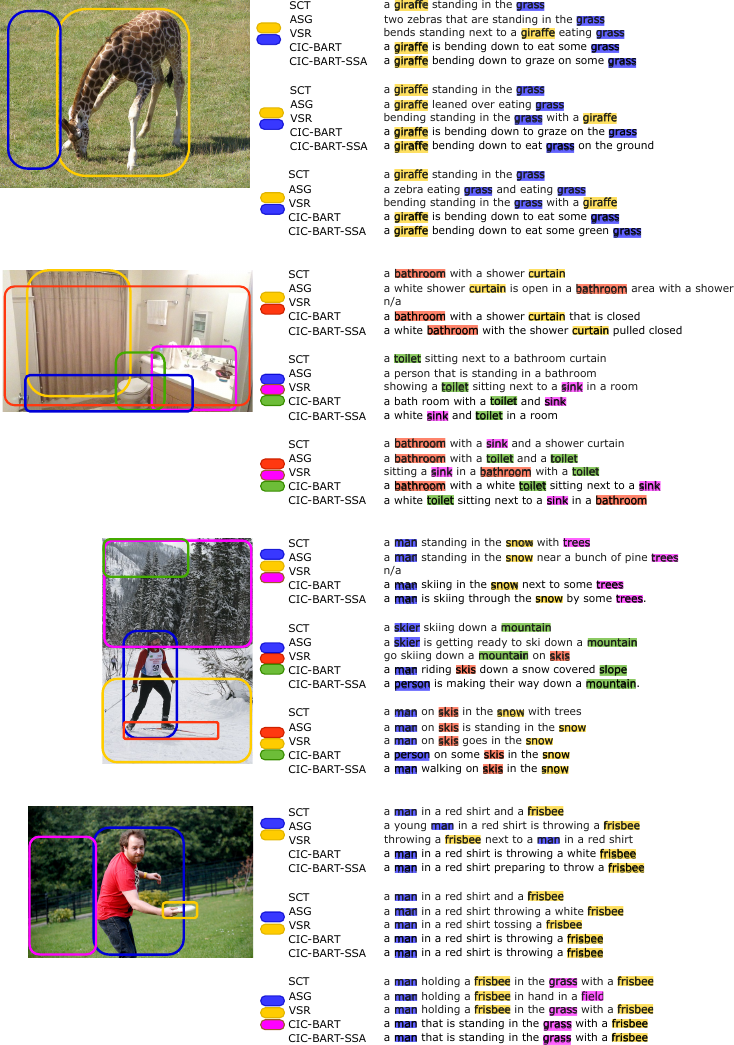}
    \caption{Qualitative examples from \cocoent test set. We include the generated controlled captions of SCT, ASG, and VSR methods and our proposed \modelname and \modelname-SSA models. \modelname was trained using \cocoent training set, whereas \modelname-SSA used \cocoentssa training set. 
    Images \url{http://images.cocodataset.org/train2017/000000001448.jpg}, \url{http://images.cocodataset.org/train2017/000000281019.jpg} and \url{http://images.cocodataset.org/val2017/000000325991.jpg} are licensed under a \href{https://creativecommons.org/licenses/by-sa/2.0/}{CC BY-SA 2.0} and image \url{http://images.cocodataset.org/val2017/000000038210.jpg} under a \href{https://creativecommons.org/licenses/by/2.0/}{CC BY 2.0}.
    }
    \label{fig:coco-qual-supp}
\end{figure}

\begin{figure}[t!]
    \centering
    \includegraphics[width=.85\linewidth]{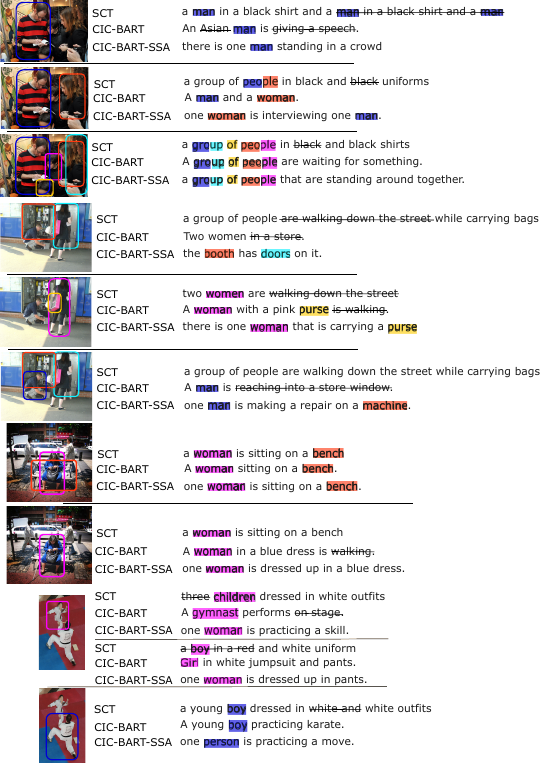}
    \caption{Qualitative examples from our SSA test set constructed from \flickrent. \modelname was trained using \flickrent training set, whereas \modelname-SSA used \flickrssa training set. Images \url{https://www.flickr.com/photos/thunderchild5/183647966/}, \url{flickr.com/photo.gne?id=151970521} and \url{flickr.com/photo.gne?id=101362133} are licensed under a Creative Commons CC BY 2.0 \url{https://creativecommons.org/licenses/by/2.0/}, and image \url{flickr.com/photo.gne?id=7249763658} under a Creative Commons PDM 1.0 \url{https://creativecommons.org/publicdomain/mark/1.0/}.}
    \label{fig:flickr-ssa-qual-supp}
\end{figure}

\begin{figure*}[t!]
    \centering
    \includegraphics[width=\linewidth]{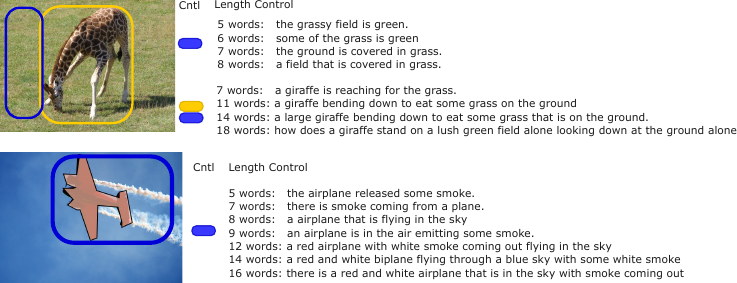}
    \caption{Qualitative examples for various length control values given a specific image sub-region. The generated controllable captions are from our trained \modelname-SSA model. Image
\url{http://images.cocodataset.org/train2017/000000001448.jpg} is licensed under Creative Commons CC BY-SA 2.0 \url{https://creativecommons.org/licenses/by-sa/2.0/} and \url{http://images.cocodataset.org/train2017/000000115178.jpg} under a Creative Commons CC BY 2.0 \url{https://creativecommons.org/licenses/by/2.0/}.}
    \label{fig:len-cntl-supp}
\end{figure*}

\subsection{Qualitative Results}\label{sec:supp-qualresults}
In \cref{fig:qual-main-supp}, we present qualitative examples from \cocoent and \flickrent test sets. In these examples, the control signals are extracted from the ground-truth captions. 
Each colored oval under `Cntl' corresponds to a bounding box of the same color in the image. The collection of ovals identifies the entities of interest, that is, the control signal. (Note that we do not show the colored ovals for image (c), since both sets of control signals include all bounding boxes.) For example, in (a), the first control signal (at the top) focuses on the regions \textit{restaurant}, \textit{man}, and \textit{food}, while the second control signal also includes the entity \textit{table}.
Each highlighted word in the generated controlled captions corresponds to the control entity of the same color, showing the match between the generated captions and the control signal.

We notice that our models outperform previous SOTA by substantially improving the quality of the generated controlled captions. This behavior was expected from our quantitative analysis, showing that our models have significantly higher text quality (G) performance. In addition, more evidently in figures (b) and (d), our models have better content controllability performance by correctly referring to all entities of interest in the control signal. Especially in the highly challenging, complex scene (d) in which many objects are present, it successfully describes all entities of interest in the generated controllable caption. 

Our \modelname-SSA model generates captions that are faithful to the control signal and better understand the relationships connecting the entities of interest. For example, in (a), it correctly identifies that the person is photographing his food rather than eating it and that the image is not black and white, or in (d) that the woman holds the purse and not the man in the background.  

In \cref{fig:coco-qual-supp} we present additional qualitative examples from \cocoent test set. We include the generated controllable captions from the baseline models (SCT, ASG, and VSR) and the proposed models \modelname and \modelname-SSA. Our qualitative examples also show that our models generate diverse, high-text-quality captions with improved content controllability when compared to the baseline models.

Next in \cref{fig:flickr-ssa-qual-supp} we present examples using the SSA augmentations control signals which are derived from \flickrent test set. We notice that our \modelname-SSA better conveys the image concept without hallucinating. For example, in the second image, it correctly describes that the man is fixing the ticket booth or that the woman carries a bag and is not walking.

Further, we conducted an experiment to evaluate our length control performance qualitatively. In the experiment, we generated controllable captions for a fixed image region and various caption length controls. We present some of our qualitative results in \cref{fig:len-cntl-supp} showing that the generated captions were faithful to the length control signal, indicating that our model is effective in controlling the length of captions. Furthermore, we observe that our model generates a diverse set of captions for a specific image region.

\subsection{
SSA vs Other Augmentation Strategies}
\label{sec:sup-augmentations-ablation}

\subsubsection{Augmentations via LLM Paraphrasing}
\label{sec:sup-llm-paraphrasing}

To understand the impact of our SSA enhancements, we perform an experiment in which we augment the original training data with paraphrases generated using an LLM, Llama-2~\cite{touvron2023llama}. We generate one paraphrase per original caption, effectively doubling the size of the training data\footnote{For 20\% of the captions, Llama generates paraphrases identical to original sentences.}.  Specifically, we instructed the Llama-2 model to rephrase the initial captions using few shot prompting like 
\begin{quote}
    \textit{If the phrase `Children wearing team uniforms playing soccer in a grassy field' can be paraphrased as `Kids in a grassy field playing soccer in uniforms', and the phrase `A little girl sitting in the middle of a restaurant and smiling for picture'  can be paraphrased as `A smiling little girl taking a picture while sitting in a restaurant', then the phrase `\{caption\}' can be paraphrased as \ldots}
\end{quote}
We replace \{caption\} with original dataset captions, relying on Llama-2 to paraphrase them. Since we only paraphrased the original sentence, we can assume that it pertains to the same set of bounding boxes since it refers to the same visual entities of interest, which is the only information required for our \modelname model.

The results in \cref{tab:ablation-table-paraphrases} (bottom panel) demonstrate that \modelname-SSA outperforms \modelname-par on all controllable captioning metrics.
We conclude that the improved performance of \modelname-SSA is not just due to the increase in training data size; the model benefits from the intricate structured and visually grounded guidance of our SSA.

\begin{table}[ht]
\caption{Performance of our SSA augmentations (\modelname-SSA) vs LLM paraphrases (\modelname-par). All models are evaluated on the original \cocoent test sets.
}
\centering
\scriptsize
\setlength{\tabcolsep}{4pt}
\begin{tabular}{@{}lcccccccc}
\toprule
Model                             & $H\uparrow$ & IoU$\uparrow$ & Hal$\downarrow$ & G$\uparrow$ & sC$\uparrow$ & D-1$\uparrow$ & D-2$\uparrow$  & L$\downarrow$\\
\midrule
\modelname-par                    & 74.3        & 76.2          & 18.8            & 72.0        & 74.9         & 36.1          & 52.7           & .19      \\
\modelname-SSA                    & \bf{78.3}   & \bf{77.2}     & \bf{17.8}       & \bf{74.8}   & \bf{82.5}    & \bf{44.6}     & \bf{63.2}      & \bf{.11} \\
\bottomrule
\end{tabular}
\label{tab:ablation-table-paraphrases}
\end{table}

In \cref{fig:sup-coco-101310,fig:sup-coco-101155,fig:sup-coco-101194}, we present some examples of the Llama-2 paraphrases of the original COCO captions. For example, the paraphrase of 1-O) is 1-Llama-2), the paraphrase of 2-O) is 2-Llama-2), and so on. We have included the SSA-generated focused captions for each image, along with the corresponding synthetic caption and its GRUEN score. SSA uses this metric to filter out poor-quality sentences (in our experiments, we set the GRUEN threshold to 0.7).

\subsubsection{Scene Graphs with LLM Paraphrasing Augmentations}
\label{sec:sup-scenegraph-augmentations}
\begin{figure}
    \centering
    \includegraphics[width=\linewidth]{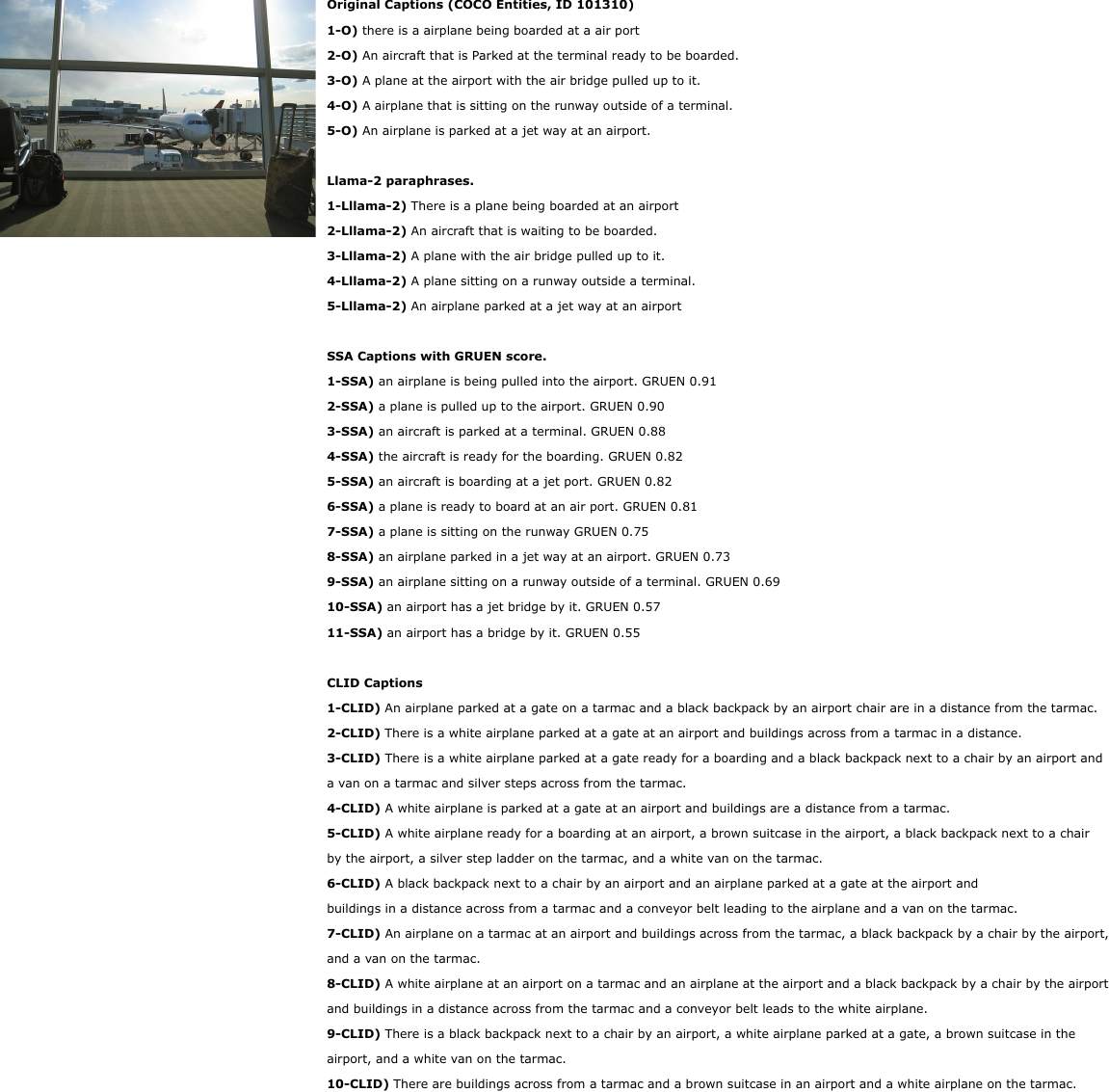}
    \caption{Qualitative examples from different augmentation strategies. We present the original five dataset captions, their Llama-2 paraphrases, our SSA (AMR-based) generated descriptions, and finally, CLID (scene-graph-based) augmentations. The GRUEN score is included for the SSA captions, which is used to filter out poorly generated sentences. Image \url{http://images.cocodataset.org/train2017/000000101310.jpg} is licensed under a Commons Creative CC BY-SA 2.0 \url{https://creativecommons.org/licenses/by-sa/2.0/}.}
    \label{fig:sup-coco-101310}
\end{figure}

\begin{figure}
    \centering
    \includegraphics[width=\linewidth]{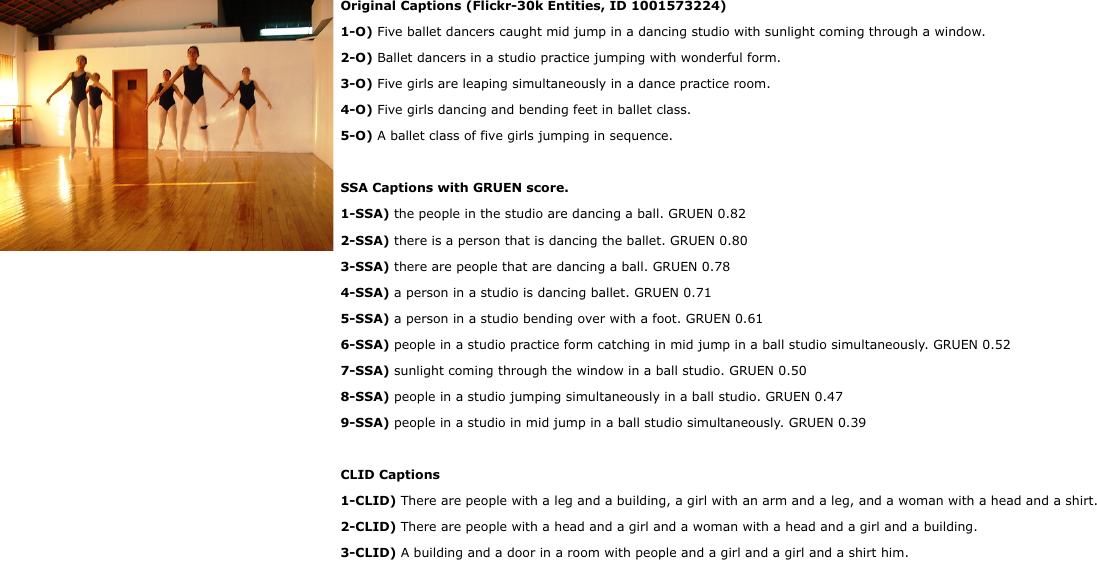}
    \caption{Qualitative examples from different augmentation strategies. We present the original five dataset captions, our SSA (AMR-based) generated descriptions, and finally, CLID (scene-graph-based) augmentations. The GRUEN score is included for the SSA captions, which is used to filter out poorly generated sentences. Image \url{https://www.flickr.com/photos/bombarosa/1001573224/} is under Creative Commons CC BY-ND 2.0 \url{https://creativecommons.org/licenses/by-nd/2.0/} license.}
    \label{fig:sup-flickr-1001573224}
\end{figure}

\begin{figure}
    \centering
    \includegraphics[width=\linewidth]{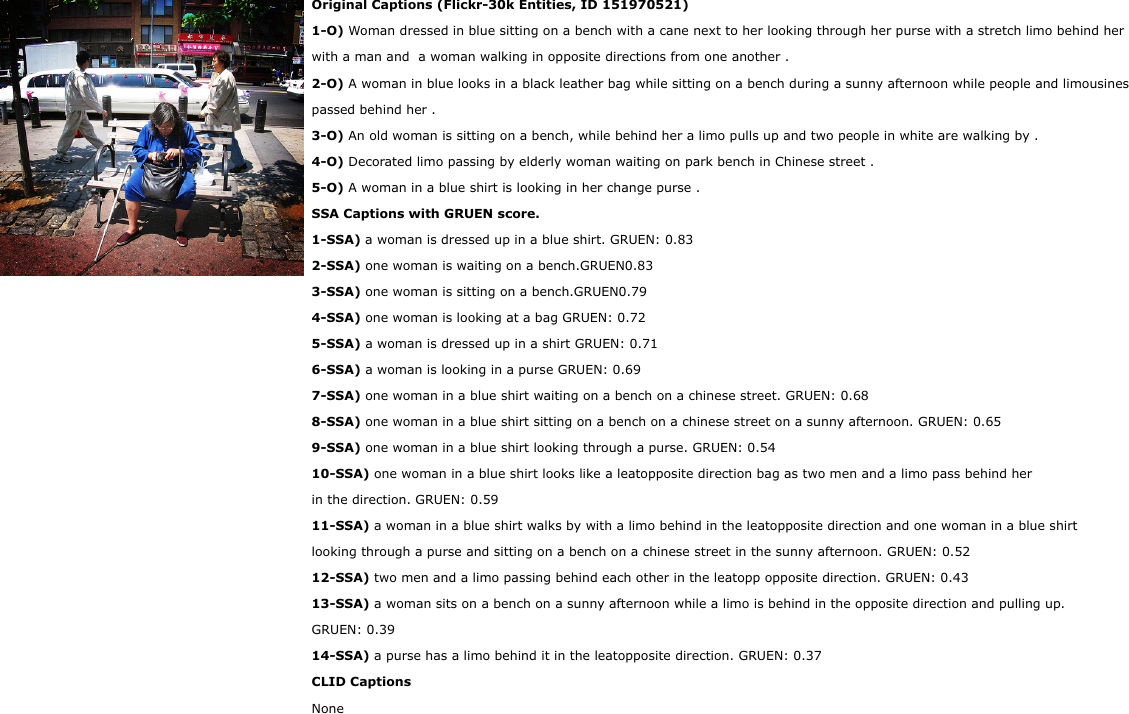}
    \caption{Qualitative examples from different augmentation strategies. We present the original five dataset captions and SSA (AMR-based) generated descriptions. The GRUEN score is included for the SSA captions, which is used to filter out poorly generated sentences. The CLID dataset does not provide additional annotations for this image. Image \url{flickr.com/photo.gne?id=151970521} is under a Creative Commons CC BY 2.0 \url{https://creativecommons.org/licenses/by/2.0/} license.}
    \label{fig:sup-flickr-151970521}
\end{figure}

The scope of this section is to illustrate the benefits of AMRs when compared to scene graphs for CIC augmentation. \cref{fig:sup-coco-101310,fig:sup-flickr-1001573224,fig:sup-flickr-151970521} depict examples from COCO-Ent and Flickr-Ent that contrast Original Captions, LLM-paraphrased captions, SSA-augmented captions, and CLID-augmented captions.

\begin{figure}
    \centering
    \includegraphics[width=\linewidth]{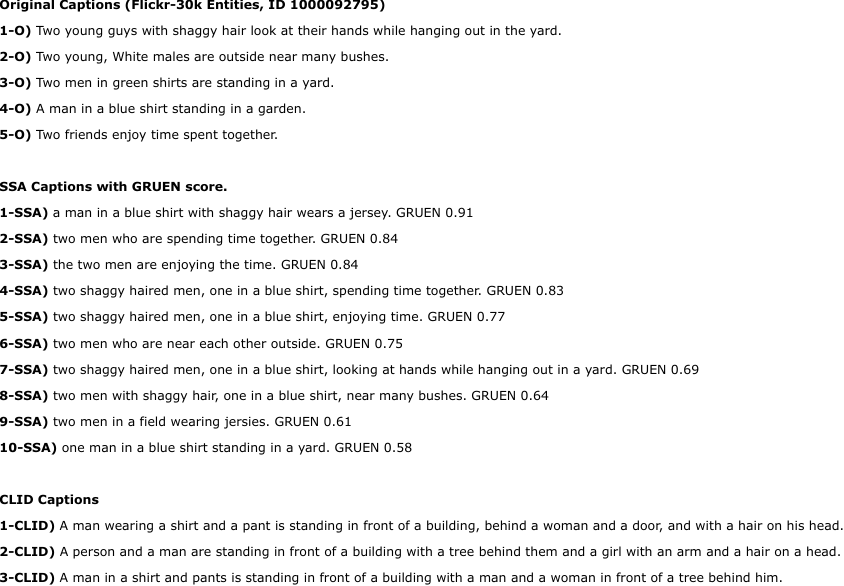}
    \caption{Qualitative examples from different augmentation strategies. We present the original five dataset captions, our SSA (AMR-based) generated descriptions, and finally, CLID (scene-graph-based) augmentations. For the SSA captions, their GRUEN score is included, which is used to filter out poorly generated sentences. \textit{The captions are from the image with ID 1000092795 from the \flickrent dataset. The image is not depicted here due to license limitations.}}
    \label{fig:sup-flickr-1000092795}
\end{figure}

\paragraph{Nature of captured relations and entities in AMR vs.~Scene Graph representations.} As stated in our main paper, prior analysis \cite{abdelsalam2022visual,zellers2018neural} has shown that existing scene graph annotations focus mainly on geometric and possessive relations. For example, in \cref{fig:sup-flickr-1000092795} examples of geometric relations are `a man in front of a door', `one man next to another man,' etc., and possessive `a man has hair,' `the man has a head' etc. Regarding entities, scene graphs focus mainly on object/body parts (hair, head, arm, etc.) and clothing (dress, shirt, etc.). On the contrary, the AMRs derived from the image captions contain a wide range of semantic relations that are inherited from the natural language image descriptions drawn from the image captioning datasets. For example, in \cref{fig:sup-flickr-1000092795}, in the original captions, we can find the semantic relations, 1-O) `the men are \textit{hanging out} in the yard', 5-O) `the friends \textit{enjoy time spent} together' which will be inherited in their AMR representations and therefore in the SSA samples (i.e. 2-SSA), 3-SSA), 5-SSA), 7-SSA)).

\paragraph{Limitations of Scene Graph representations.} Scene graphs are useful in capturing the visual elements of a scene and their relationships, such as geometric and possessive relationships. However, they lack the ability to represent abstract concepts like time-related information, such as `a sunny morning' or `a quiet afternoon'. For instance, in \cref{fig:sup-flickr-151970521} caption 2-O), the phrase `during a sunny afternoon' is not directly related to low-level semantics like individual visual objects and their relationships. Rather, it relates to higher-level reasoning, such as observing the shadows of the trees and the annotators' experiences that helped them conclude that the picture was taken during a sunny afternoon.

Another important difference between AMRs and scene graphs is that the edges of an AMR carry linguistic information, which is not the case with scene graphs. In \cref{fig:amr-supp}, we can see only a fraction of the available edge roles, which are crucial when converting vgAMRs to natural language sentences as they help generate accurate descriptions. On the other hand, scene graph edges lack the ability to convey additional linguistic information, except for the edge direction, which indicates the object and subject of a relation. This lack of additional information makes it challenging to augment captions using sampled scene graphs and forces reliance on LLM paraphrasing, which, as previously discussed in \cref{sec:sup-llm-paraphrasing}, can introduce errors and inaccuracies (increased hallucinations (Hal) and poorer text quality (G) as seen in \cref{tab:ablation-table-paraphrases}). For instance, in the scene graph-based augmentation of CLID \cite{hirsch2022clid}, they first place the sampled scene graph node names in a sequence and then ask an LLM paraphraser to generate a sentence.

\begin{figure}
    \centering
    \includegraphics[width=.7\linewidth]{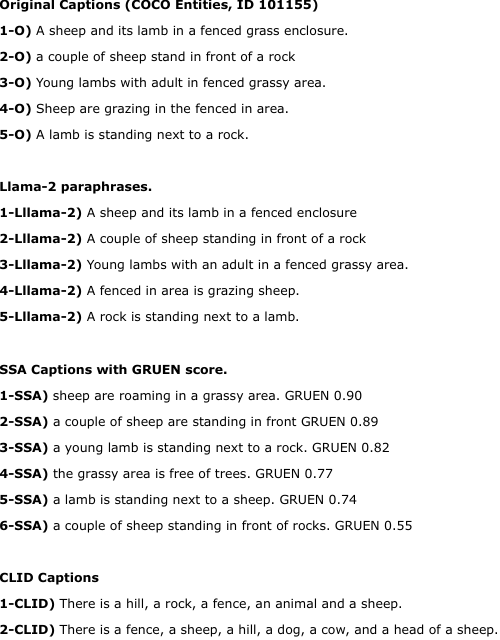}
    \caption{Qualitative examples from different augmentation strategies. We present the original five dataset captions, their Llama-2 paraphrases, our SSA (AMR-based) generated descriptions, and finally, CLID (scene-graph-based) augmentations. For the SSA captions, their GRUEN score is included, which is used to filter out poorly generated sentences. \textit{The captions are from the image with ID 101155 from the \cocoent dataset. The image is not depicted here due to license limitations.}}
    \label{fig:sup-coco-101155}
\end{figure}

\begin{figure}
    \centering
    \includegraphics[width=\linewidth]{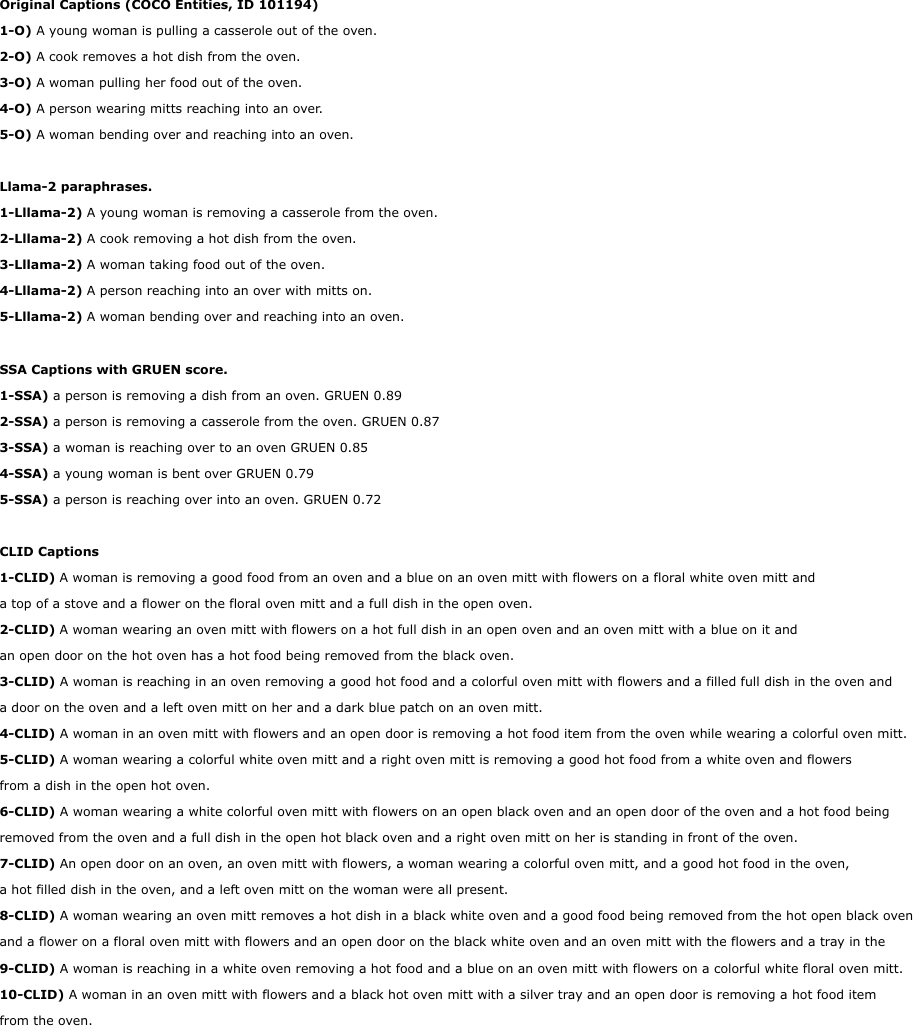}
    \caption{Qualitative examples from different augmentation strategies. We present the original five dataset captions, their Llama-2 paraphrases, our SSA (AMR-based) generated descriptions, and finally, CLID (scene-graph-based) augmentations. For the SSA captions, their GRUEN score is included, which is used to filter out poorly generated sentences. \textit{The captions are from the image with ID 101194 from the \cocoent dataset. The image is not depicted here due to license limitations.}}
    \label{fig:sup-coco-101194}
\end{figure}

\begin{figure}
    \centering
    \includegraphics[width=\linewidth]{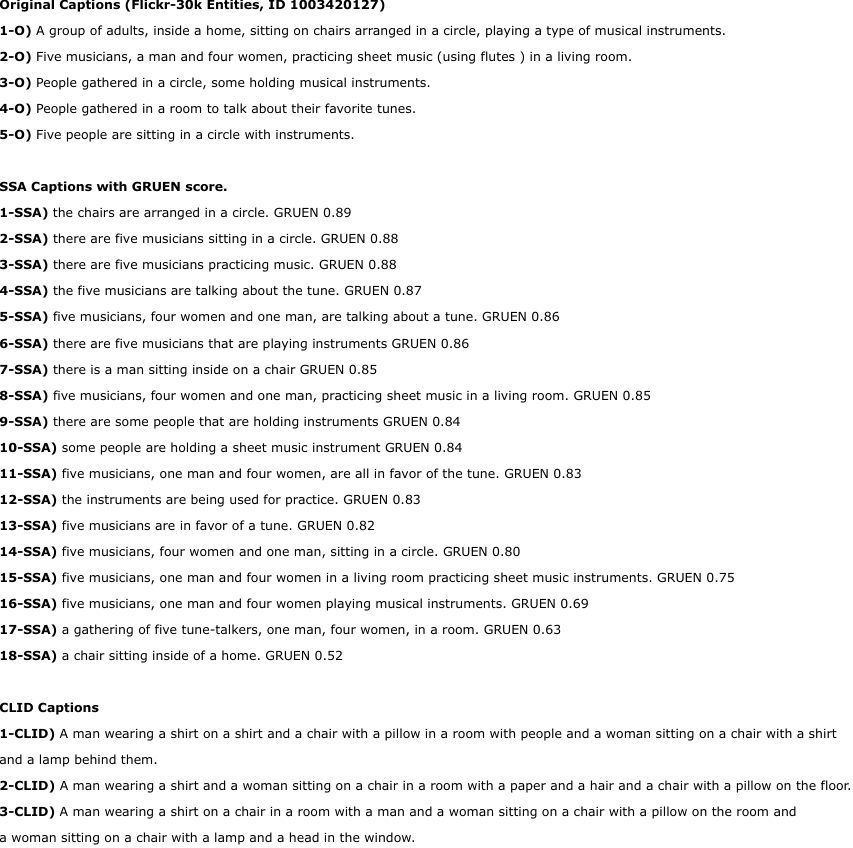}
    \caption{Qualitative examples from different augmentation strategies. We present the original five dataset captions, our SSA (AMR-based) generated descriptions, and finally, CLID (scene-graph-based) augmentations. For the SSA captions, their GRUEN score is included, which is used to filter out poorly generated sentences. \textit{The captions are from the image with ID 1003420127 from the \flickrent dataset. The image is not depicted here due to license limitations.}}
    \label{fig:sup-flickr-1003420127}
\end{figure}

In \cref{fig:sup-coco-101155,fig:sup-coco-101194,fig:sup-coco-101310,fig:sup-flickr-151970521,fig:sup-flickr-1000092795,fig:sup-flickr-1001573224,fig:sup-flickr-1003420127}, we can see instances from our SSA (AMR-based) and CLID (scene graph-based) augmentations for a particular image. Augmentations in the CLID dataset are not visually grounded and, hence, cannot be utilized for spatial CIC. However, our SSA approach includes visual grounding information, which enables it to effectively generate visually grounded augmentations, making them suitable for spatial CIC tasks as well. 

CLID augmentations mainly describe the geometric and possessive relationships of the visual entities, and not their semantic relations. The main focus of these augmentations is on body parts (i.e. in \cref{fig:sup-flickr-1001573224}, caption 1-CLID: `people with a leg', `a girl with an arm and a leg', and `a woman with a head') as well as clothing. This is expected, since this type of information is typically captured in existing scene graphs. It is also noticed that most of the generated sentences are difficult to read, containing redundancies and hallucinations. This may be the result of scene graph annotation or generation errors or due to LLM-induced hallucinations and poor quality text generation.
On the other hand, the SSA augmentations are based on AMR representations and capture various types of relations, including semantic, geometric, possessive, etc. Unlike other methods, they provide a more natural and human-like description of visual entities, as they are derived from the original dataset of human-annotated captions.

\end{document}